\documentclass[10pt,twocolumn]{article}
\usepackage{graphicx}
\usepackage{amsmath,amssymb,times,cite}
\usepackage{booktabs}
\usepackage[dvipsnames]{xcolor}

\newcommand{\op}[1]{\operatorname{#1}}

\newcommand{\bm}[1]{\mathbf{#1}} 

\newcommand\T{{\mathpalette\raiseT\intercal}}
\newcommand\raiseT[2]{%
\setbox0\hbox{$#1{#2}$}\raise\dp0\box0}

\title{\Large\textbf{Multi-hop Graph Transformer Network for 3D Human Pose Estimation}}
\author{Zaedul Islam and A. Ben Hamza\\
Concordia Institute for Information Systems Engineering\\
Concordia University, Montreal, QC, Canada
}
\date{}

\begin{document}
\maketitle

\begin{abstract}
Accurate 3D human pose estimation is a challenging task due to occlusion and depth ambiguity. In this paper, we introduce a multi-hop graph transformer network designed for 2D-to-3D human pose estimation in videos by leveraging the strengths of multi-head self-attention and multi-hop graph convolutional networks with disentangled neighborhoods to capture spatio-temporal dependencies and handle long-range interactions. The proposed network architecture consists of a graph attention block composed of stacked layers of multi-head self-attention and graph convolution with learnable adjacency matrix, and a multi-hop graph convolutional block comprised of multi-hop convolutional and dilated convolutional layers. The combination of multi-head self-attention and multi-hop graph convolutional layers enables the model to capture both local and global dependencies, while the integration of dilated convolutional layers enhances the model's ability to handle spatial details required for accurate localization of the human body joints. Extensive experiments demonstrate the effectiveness and generalization ability of our model, achieving competitive performance on benchmark datasets.
\end{abstract}


\section{Introduction}
3D human pose estimation is a fundamental and challenging task in computer vision, with a myriad of applications spanning action recognition~\cite{Song2021Survey,Luvizon2021Action}, autonomous driving~\cite{Zanfir2023Auto}, sports analytics~\cite{Ingwerse2032Sport}, and healthcare diagnostics~\cite{Gu2019Therapy}. At its core, it involves the prediction of 3D coordinates of human body joints from images or videos with the goal of understanding human movements and interactions. With the advent of deep learning, several deep neural networks~\cite{zhou2016deep,park20163d,sun2018integral,pavlakos2017coarse,sun2017compositional,yang20183d,chen2020towards,
lee2018propagating,chen20173d,tome2017lifting,tekin2017learning} have been devised for 3D human pose estimation, classified into one-stage and two-stage approaches. One-stage methods aim to directly regress the 3D pose from input images, whereas two-stage methods first predict intermediate representations, such as 2D joint locations, using 2D pose detectors before lifting them to 3D space. Two-stage methods typically exhibit better performance compared to one-stage approaches, particularly when combined with robust 2D joint detectors~\cite{chen2018cascaded,sun2019deep}, as they improve the accuracy of 3D pose estimation while addressing depth ambiguity challenges. These methods typically employ a 2D pose detector in the first stage, followed by a lifting network for predicting the 3D pose locations from the 2D predictions in the second stage. Our proposed framework falls under the category of two-stage methods. Despite notable progress~\cite{Zheng2023Survey}, several challenges persist in 3D human pose estimation. Self-occlusions often occur when body parts obscure each other, making it difficult for the model to accurately estimate the position of occluded joints. Moreover, depth ambiguity arises due to variations in body shape, occlusions, and self-occlusions, leading to multiple 3D pose possibilities for the same 2D projections. Overcoming these challenges remains critical for improving the robustness and accuracy of 3D human pose estimation methods.

In recent years, graph convolutional network (GCN)-based methods~\cite{zhao2019semantic,azizi20223d,zhang2022group} and approaches built on the Transformer architecture~\cite{zhao2022graformer,PoseFormer:2021} have become more prevalent in 3D human pose estimation. While effective, GCN-based methods are limited in their ability to capture dependencies between body joints that are beyond immediate neighbors. The standard GCN architecture, which relies on the adjacency matrix to propagate information, considers only the direct connections between nodes in the graph, resulting in a relatively local view of the graph structure. As a consequence, these methods may not fully exploit the long-range interactions and complex dependencies that exist in human body movements. To address this challenge, various approaches~\cite{zou2020high,quan2021higher} employ high-order graph convolutions, which leverage higher powers of the adjacency matrix, thereby allowing information to be propagated through multiple hops in the graph. By using higher powers of the adjacency matrix, the model can gather information from not only the immediate neighbors but also nodes that are further away in the graph, enhancing its ability to consider global context and complex relationships between body joints. Using higher powers can, however, lead to the biased weighting problem~\cite{Liu2020Disentangle}. This issue arises due to the nature of undirected graphs, which can contain cyclic walks. In such graphs, edge weights can be influenced by the number of hops or steps required to traverse from one node to another. As a result, the weights tend to be biased towards nodes that are closer in terms of hop count, and this bias can lead to an overemphasis on local connections while neglecting long-range dependencies. To mitigate this issue, we incorporate disentangled neighborhoods in our proposed framework so that our model can capture information from nodes that are further away without being overly influenced by local connections. This helps enhance the ability of our model to capture both short-range and long-range dependencies effectively. Also, GCN-based methods are constrained by the limited receptive fields. To address this limitation, we incorporate dilated graph convolutions into our model architecture. By adjusting the dilation rate of the convolutional kernels, our model can capture information from a wider region of the graph, effectively incorporating global contextual information into the feature representation.

On the other hand, Transformer-based methods~\cite{zhao2022graformer,PoseFormer:2021} leverage self-attention mechanisms~\cite{Vaswani2017Tranformers} to process a sequence of 2D joint locations across frames in a video, enabling them to efficiently model dependencies between different body joints in the sequence. By attending to relevant joints in the sequence, Transformers can effectively capture the temporal and spatial relationships~\cite{PoseFormer:2021} between body joints, leading to improved pose estimation accuracy. Although self-attention mechanisms enable the body joints to interact by capturing global visual information through long-range dependencies and contextual information, they often neglect the inherent graph structure information among the joints (i.e., the adjacency relation of joints). By neglecting the graph structure, Transformer-based methods might overlook the specific dependencies between joints, potentially leading to suboptimal predictions. To overcome this limitation, we combine the strengths of Transformer-based methods with GCNs by incorporating multi-hop graph convolutions with the aim of explicitly taking advantage of the graph structure information, allowing our model to better capture the relationships between body joints from various hop distances. We achieve this integration through a two-pathway design, where one pathway handles the self-attention mechanisms of the Transformer, capturing global visual information, and the other pathway performs graph convolutions to consider the graph structure information and interactions between joints.

\begin{figure}[!htb]
\begin{center}
\includegraphics[scale=.52]{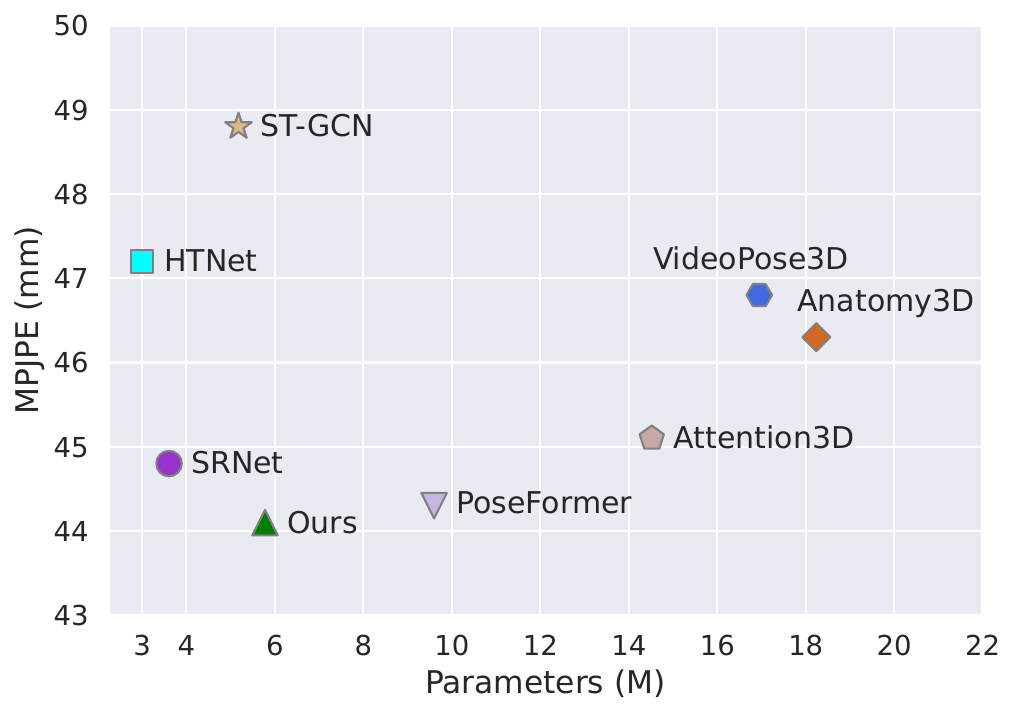}
\end{center}
\caption{Performance and model size comparison between our model and state-of-the-art temporal methods for 3D human pose estimation, including PoseFormer~\cite{PoseFormer:2021}, VideoPose3D~\cite{pavllo20193d}, ST-GCN~\cite{YujunCai:19}, SRNet~\cite{zeng2020srnet}, Attention3D~\cite{liu2020attention}, Anatomy3D~\cite{chen2021anatomy}, and HTNet~\cite{cai2023htnet}. Lower Mean Per Joint Position Error (MPJPE) values indicate better performance. Evaluation is conducted on the Human3.6M dataset using detected 2D joints as input.}
\label{Fig:Figure1}
\end{figure}

In this paper, we present a novel approach, dubbed Multi-hop Graph Transformer Network (MGT-Net), for 2D-to-3D human pose estimation in videos. The architecture of the proposed network is comprised of two main building blocks: a graph attention block and a multi-hop graph convolutional block. The graph attention block is composed of a multi-head self-attention layer that captures not only local relationships among neighboring joints but also global context information across all joints, and a graph convolutional layer with learnable adjacency matrix. By learning the adjacency matrix, our model can adaptively adjust the connections between the body joints, allowing it to capture complex dependencies in the graph. The multi-hop graph convolutional block, on the other hand, consists of multi-hop graph convolutional layers with disentangled neighborhoods that enable the model to capture dependencies between nodes at varying hop distances and dilated convolutional layers to enhance the model's receptive field without increasing the number of learnable parameters. One of the key strengths of our proposed model is its simplicity, which stands in contrast to many existing spatio-temporal approaches for 3D human pose estimation. Despite its straightforward design, our model achieves competitive performance, surpassing strong baseline methods in accuracy, as illustrated in Figure~\ref{Fig:Figure1}. This is particularly noteworthy as our model maintains a compact model size, making it computationally efficient. In summary, we make the following key contributions:

\begin{itemize}
\item We propose a multi-hop graph transformer network (MGT-Net), which leverages multi-head self-attention, multi-hop graph convolutions with disentangled neighborhoods, and dilated convolutions to effectively capture both long-range dependencies and local-global contextual information.
\item We design a network architecture composed of a graph attention block and a multi-hop graph convolutional block to capture spatial dependencies and model the complex interactions among body joints.
\item We demonstrate through extensive experiments and ablation studies the effectiveness and generalization ability of the proposed MGT-Net model, achieving competitive performance in 3D human pose estimation on two benchmark datasets while retaining a small model size.
\end{itemize}

The remainder of this paper is structured as follows. In Section 2, we discuss the related work. In Section 3, we present our proposed framework. In Section 4, we present empirical results comparing our model with state-of-the-art approaches on two standard benchmarks. Finally, we conclude in Section 5.

\section{Related Work}
Several lines of work are related to ours. In this section, we provide an overview of 3D human pose estimation methods, with a particular focus on GCN- and Transformer-based approaches.

\medskip\noindent\textbf{3D Human Pose Estimation.}\quad Recent approaches to 3D human pose estimation can be categorized into two main groups: one-stage and two-stage methods~\cite{zhou2016deep,park20163d,sun2018integral,pavlakos2017coarse,sun2017compositional,yang20183d,chen2020towards,
lee2018propagating,chen20173d,tome2017lifting,tekin2017learning}. One-stage methods, also known as direct regression methods, aim to directly predict the 3D joint locations from an input image or video without requiring any intermediate predictions. For instance, Pavlakos \textit{et al.}~\cite{pavlakos2017coarse} introduce an end-to-end approach that formulates 3D human pose estimation as a 3D keypoint localization problem in a voxel space, and also propose a coarse-to-fine prediction scheme to  handle the large dimensionality of the volumetric representation and iteratively refine the 3D joint location estimates. However, one-stage methods usually suffer from depth ambiguity, which arises because the 3D pose estimation problem is inherently underconstrained, meaning that there are multiple possible 3D poses that can explain the same 2D observations. Also, they do not perform well when dealing with complex poses or occlusions. On the other hand, two-stage methods, also known as indirect regression methods, take a different approach. They first predict intermediate representations, such as 2D joint locations, and then utilize these intermediate predictions to estimate the 3D joint locations. For example, Martinez \textit{et al.}~\cite{martinez2017simple} propose a lightweight deep neural network architecture for 2D-to-3D human pose estimation. The core of the architecture is a multilayer neural network comprised of a linear layer designed to process the input data and learn the relationships between the 2D joint detections and the corresponding 3D human pose, followed by batch normalization, an activation function, a dropout layer, and residual connections that enable the gradient flow through the network during training. Two-stage methods tend to be more accurate, especially when combined with robust 2D joint detectors, and our proposed approach falls under this category.

\medskip\noindent\textbf{GCN-based Methods.}\quad In recent years, GCNs have emerged as a powerful approach for 3D human pose estimation, leveraging the inherent graph structure of human skeletons to capture spatial dependencies between body joints. Early works primarily focused on exploiting graph representations to model human body structures. This arises from the fact that the human skeleton can naturally be represented as a graph, with nodes corresponding to joints and edges capturing the connections between the joints. Zhao \textit{et al.}~\cite{zhao2019semantic} introduce SemGCN, a semantic graph convolutional network that explores both semantic information and relationships within the graph using a semantic graph convolution (SemGConv). By learning weights for edges in a channel-wise manner, SemGConv captures both local and global relationships among nodes, which are further enhanced by interleaving with non-local layers. Azizi \textit{et al.}~\cite{azizi20223d} encode the local transformation between the joints using the M\"{o}obius transformation applied to the eigenvalues of the Laplacian matrix. Zhang~\cite{zhang2022group} extended the concept of group convolution in convolutional neural networks to GCNs, showing that decoupling the aggregation mechanism in GCNs helps improve the performance of the graph convolution. Zou \textit{et al.}~\cite{zou2021modulated} introduce a modulated GCN comprised of weight modulation and affinity modulation. The weight modulation exploits different modulation vectors for different nodes, which enables independent feature transformations. Affinity modulation, on the other hand, explores additional joint correlations beyond those defined by the human skeleton. However, these methods only consider 1-hop neighbors while neglecting valuable multi-scale contextual information. To address this issue, Zou \textit{et al.}~\cite{zou2020high} design a high-order graph convolutional network to capture dependencies that exist between body joints, which takes into consideration neighboring joints that are multiple hops away. Lee \textit{et al.}~\cite{multihop2022}, propose a multi-hop modulated graph convolutional network, which models a wider range of interactions between body parts by using unique adjacency matrices for various hop distances and modulating features of nodes to capture long-range dependencies. In~\cite{Zaedul2023GSNet}, Islam \textit{et al.} introduced a Gauss-Seidel graph neural network (GS-Net) that leverages graph neural networks and spectral graph filtering, emphasizing smoothness in the filtered feature vectors of neighboring graph nodes. GS-Net focuses on spatial modeling rather than explicitly incorporating temporal modeling components. The spatial relationships between the body joints are modeled using a GCN-based architecture that utilizes an iterative layer-wise propagation rule, and spectral graph filtering is applied for spatial processing. In other words, GS-Net's design and focus make it a spatial method for 3D human pose estimation, and its primary strength lies in spatial graph-based processing. Our approach differs from these GCN-based methods in several aspects. We employ multi-hop graph convolutional operations and multi-head self-attention, which allow our model to effectively capture both spatial and temporal dependencies, enabling it to gather long-range information and consider global context, and hence resulting in stronger representation power. We also employ a graph convolutional layer with a trainable adjacency matrix, which allows our model to adaptively learn the graph structure based on the input data, resulting in a more flexible and data-driven representation of the relationships between body joints. Moreover, we leverage dilated convolutional layers to enhance the model's receptive field without increasing the number of learnable parameters. By design, our network adopts skip connections to learn both high-level and low-level features simultaneously, contributing to improved overall performance of our approach while retaining a compact model size.
	
\medskip\noindent\textbf{Transformer-based Methods.}\quad Inspired by the success of Transformers in natural language processing and computer vision tasks~\cite{Vaswani2017Tranformers,Dosovitskiy2021VT}, Transformer-based methods for 3D human pose estimation have recently emerged as a promising approach for capturing long-range dependencies and global context between body joints in video sequences by leveraging self-attention mechanisms. Self-attention allows each body joint to interact with all other joints in the sequence, capturing their relative importance and relevance. Zhao \textit{et al.}~\cite{zhao2022graformer} present GraFormer, a model built on the Transformer architecture to model the relations between the different joints. It is comprised of a graph attention module and a Chebyshev graph convolutional module. However, Chebyshev graph convolutions approximate the spectral graph convolution by truncating the Chebyshev polynomial expansion. This approximation introduces errors, which can lead to the mixing of information from distant body joints into local joint features. Such errors can negatively impact the accuracy of pose estimation, especially for complex poses. Zheng \textit{et al.}~\cite{PoseFormer:2021} introduce PoseFormer, a Transformer-based model composed of a spatial transformer module that employs spatial self-attention layers, considering the positional information of 2D joints, to generate a latent feature representation for each frame, and a temporal transformer module that analyzes global dependencies across frames. Since PoseFormer is a pure Transformer-based model that directly models the spatial and temporal aspects of the input video sequence, it does not explicitly consider the graph structure of the human body, and hence it may not fully exploit the inherent graph relationships that exist between body joints, potentially leading to suboptimal performance. More recently, Cai \textit{et al.} propose HTNet~\cite{cai2023htnet}, a human topology aware network comprised of a local joint-level connection module based on GCNs to model physical connections between adjacent joints at the joint level, an intra-part constraint module to provide constraints for intra-part joints at the part level, and a global body-level interaction module based on multi-head self-attention to extract global features among inter-part joints at the body level. However, one key limitation of HTNet is the challenge in designing an optimal connection structure for its three modules. Also, the series structure, where the modules are connected sequentially, can increase the model size.

\medskip {In comparison to the aforementioned methods, our approach, MGT-Net, combines the strengths of multi-head self-attention and multi-hop graph convolutions with disentangled neighborhoods. By leveraging both local and global contextual information and capturing long-range dependencies effectively, MGT-Net aims to improve the accuracy and robustness of 3D human pose estimation in challenging scenarios. Moreover, we integrate dilated graph convolutions into our network architecture to enhance the model's receptive field, enabling it to capture larger contextual information and better understand the dependencies between body joints across different scales and distances.

\section{Proposed Method}
\subsection{Preliminaries and Problem Statement}
Consider an attributed graph $\mathcal{G}=(\mathcal{V},\mathcal{E}, \bm{X})$, where $\mathcal{V}=\{1,\ldots,N\}$ is a set of nodes that correspond to body joints, $\mathcal{E}$ is the set of edges representing connections between two neighboring body joints, and $\bm{X}$ is an $N\times F$ feature matrix of node attributes whose $i$-th row is an $F$-dimensional feature vector associated to node $i$. The graph structure is encoded by an $N\times N$ adjacency matrix $\bm{A}$ whose $(i,j)$-th entry is equal to 1 if there the edge between neighboring nodes $i$ and $j$, and 0 otherwise. We denote by $\hat{\bm{A}}=\tilde{\bm{D}}^{-\frac{1}{2}}\tilde{\bm{A}}\tilde{\bm{D}}^{-\frac{1}{2}}$ the normalized adjacency matrix with self-added loops, where $\tilde{\bm{A}}=\bm{A}+\bm{I}$, $\bm{I}$ is the identity matrix, $\tilde{\bm{D}}=\op{diag}(\tilde{\bm{A}}\bm{1})$ is the diagonal degree matrix, and $\bm{1}$ is an $N$-dimensional vector of all ones.

\medskip\noindent\textbf{Problem Formulation.}\quad Let $\mathcal{D}=\left\{\left(\mathbf{x}_{i}, \mathbf{y}_{i}\right)\right\}_{i=1}^{N}$ be a training set consisting of 2D joint positions $\bm{x}_{i}\in\mathcal{X}\subset\mathbb{R}^2$ and their associated ground-truth 3D joint positions $\bm{y}_{i}\in\mathcal{Y}\subset\mathbb{R}^3$. The aim of 3D human pose estimation is to learn the parameters $\bm{w}$ of a regression model $f: \mathcal{X} \rightarrow \mathcal{Y}$ by finding a minimizer of the following loss function
\begin{equation}	
\bm{w}^{*}=\arg\min_{\bm{w}}\frac{1}{N}\sum_{i=1}^{N}l(f(\bm{x}_{i}),\bm{y}_{i}),
\end{equation}
where $l(f(\bm{x}_{i}),\bm{y}_{i})$ is an empirical regression loss function.

\subsection{Multi-hop Graph Convolutional Networks}
\medskip\noindent\textbf{Revisiting High-order GCNs.}\quad GCNs and their variants have emerged as a promising approach for 3D human pose estimation, addressing the challenges of modeling complex spatial relationships and capturing contextual dependencies between body joints. By representing the human skeleton as a graph and leveraging graph convolution operations, GCNs can effectively learn joint interactions and infer accurate 3D pose estimations. Given an input feature matrix $\bm{H}^{(\ell)}\in\mathbb{R}^{N\times F_{\ell}}$ of the $\ell$-th layer with $F_{\ell}$ feature maps, the output feature matrix $\bm{H}^{(\ell+1)}$ of GCN is obtained by applying the following layer-wise propagation rule:
\begin{equation}
\bm{H}^{(\ell+1)}=\sigma(\hat{\bm{A}}\bm{H}^{(\ell)}\bm{W}^{(\ell)}),\quad \ell=0,\dots,L-1,
\label{Eq:GCNProp}
\end{equation}
where $\hat{\bm{A}}$ is the normalized adjacency matrix with self-added loops, $\bm{W}^{(\ell)}\in\mathbb{R}^{F_{\ell}\times F_{\ell+1}}$ is a trainable weight matrix with $F_{\ell +1}$ feature maps, and $L$ is the number of layers. An activation function $\sigma(\cdot)$ such as $\text{ReLU}(\cdot)=\max(0,\cdot)$ is applied to introduce non-linearity to the aggregated features. The input of the first layer is the initial feature matrix $\bm{H}^{(0)}=\bm{X}$. By applying the propagation rule iteratively for multiple layers, the GCN can capture and propagate information through the graph, enabling it to learn meaningful node representations.

\medskip\noindent The aggregation scheme of GCN is limited to using immediate neighboring nodes as illustrated in Figure~\ref{Fig:Skeleton} (left), which means it only considers direct connections between nodes. Hence, it fails to capture long-range dependencies that may exist between nodes that are further apart in the graph. To address this limitation, high-order GCNs are often employed~\cite{zou2020high,quan2021higher}, as they extend the aggregation scheme to $K$-hop neighbors, allowing for the capture of long-range dependencies between nodes in the graph via the following update rule:
\begin{equation}
\bm{H}^{(\ell+1)}=\sigma\left(\sum_{k=0}^{K}\hat{\bm{A}}^{k}{\bm{H}}^{(\ell)}\bm{W}^{(\ell)}_{k}\right),
\label{Eq:HO}
\end{equation}
where $\hat{\bm{A}}^{k}$ is the $k$-th power of the normalized adjacency matrix whose $(i,j)$-th element counts the number of walks of length $k$ between nodes $i$ and $j$, and $\bm{W}^{(\ell)}_{k}$ is a learnable weight matrix associated with $\hat{\bm{A}}^{k}$. The update rule sums the contributions from different powers of the normalized adjacency matrix, ranging from $k=0$ to $k=K$, where $K$ represents the maximum number of hops considered. This summation allows the model to capture information from neighboring nodes up to $K$ hops away. By considering multiple hops, high-order GCNs can incorporate information from more distant nodes and enhance their ability to capture global context and relationships within the graph structure.

While high-order GCNs are effective at capturing long-range dependencies, they are, however, prone to the biased weighting problem. This issue arises because undirected graphs can have cyclic walks, which can result in edge weights being biased towards closer nodes compared to further nodes, leading to preferential emphasis on local connections and neglecting long-range dependencies. Also, as the value of $k$ increases, the influence of each node's features spreads to increasingly distant neighbors. This can result in oversmoothing, where the node representations become overly similar and lose local discriminative information. In addition, computing higher powers of the normalized adjacency matrix can be computationally expensive, especially for large graphs. The computational complexity grows exponentially with the power of the normalized adjacency matrix, limiting the scalability of the approach. To address these limitations, we employ a multi-hop GCN aggregation scheme that leverages the $k$-adjacency matrix with the aim of removing redundant dependencies between node features from different neighborhoods~\cite{Liu2020Disentangle}. In contrast to the $k$-th power of the normalized adjacency matrix that focuses on the number of walks of length $k$, the $k$-adjacency matrix emphasizes the direct connections within $k$ hops, as depicted in Figure~\ref{Fig:Skeleton} (right).

\begin{figure}[!htb]
\centering
\includegraphics[scale=.23]{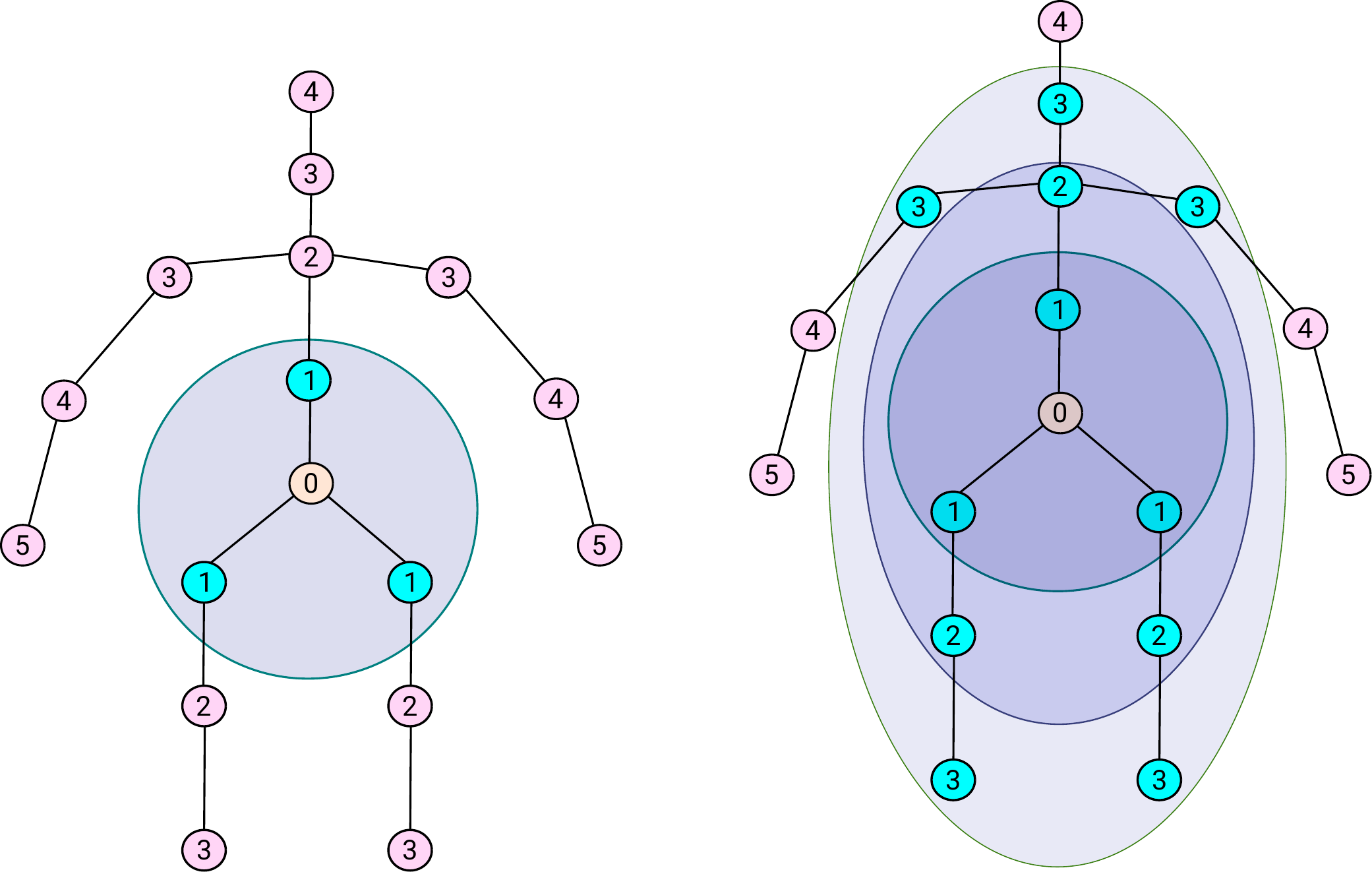}
\caption{Visual comparison between the standard graph convolution, which only considers the 1-hop neighbors, and the multi-hop graph convolution, which takes into account neighbors at different distances. The node label $k\in\{0,\dots,5\}$ indicates that the corresponding body joint is a $k$-hop neighbor of the pelvis (i.e., root node denoted by 0).}
\label{Fig:Skeleton}
\end{figure}

\medskip\noindent\textbf{Multi-hop GCNs with Disentangled Neighborhoods.}\quad  We aim to address the biased weighting problem and capture long-range dependencies more effectively. To this end, we leverage the $k$-adjacency matrix $\tilde{\bm{A}}_k$ whose $(i,j)$-th element is given by
\begin{equation}
[\tilde{\bm{A}}_{k}]_{ij}=
\begin{cases}
1 & \text{if } d_{ij}= k, \\
1 & \text{if } i = j, \\
0 & \text{otherwise},
\end{cases}
\end{equation}
where $d_{ij}$ denotes the shortest distance in the number of hops between nodes $i$ and $j$. It is important to note that incorporating self-loops in the $k$-adjacency matrix allows each joint to have a connection with itself, ensuring that the identity information of each joint is preserved. This is particularly important when dealing with nodes that do not have $k$-hop neighbors. The self-loop ensures that each joint's features are taken into account during the aggregation process, even if it does not have any direct neighbors within the specified hop distance. Note that $\tilde{\bm{A}}_{k}$ can be seen as an extension of the standard adjacency matrix $\tilde{\bm{A}}$ with self-added loops to incorporate relationships beyond immediate neighbors. While $\tilde{\bm{A}}$ captures the connections between directly connected nodes, the $k$-adjacency matrix $\tilde{\bm{A}}_k$ considers relationships up to a distance of $k$ hops, thereby capturing the relationships between neighboring joints up to a distance of $k$ hops.

Using the $k$-adjacency matrix, the layer-wise propagation rule of the multi-hop GCN can be defined as follows:
\begin{equation}
\bm{H}^{(\ell+1)}=\sigma\left(\sum_{k=0}^{K}\hat{\bm{A}}_{k}\bm{H}^{(\ell)}\hat{\bm{W}}_{k}^{(\ell)}\right),
\label{Eq:Multi-hop}
\end{equation}
where $\hat{\bm{A}}_{k}=\tilde{\bm{D}}_{k}^{-\frac{1}{2}}\tilde{\bm{A}}_{k}\tilde{\bm{D}}_{k}^{-\frac{1}{2}}$ is the normalized $k$-adjacency matrix, and $\tilde{\bm{D}}_{k}=\op{diag}(\tilde{\bm{A}}_{k}\bm{1})$ is the associated diagonal degree matrix. The multiplication $\hat{\bm{A}}_{k}\bm{H}^{(\ell)}\hat{\bm{W}}_{k}^{(\ell)}$ performs the graph convolution operation, where the normalized $k$-adjacency matrix $\hat{\bm{A}}_{k}$ is applied to the input features $\bm{H}^{(\ell)}$ and $\hat{\bm{W}}_{k}^{(\ell)}$ is is the associated weight matrix. The result is then passed through the activation function $\sigma(\cdot)$ to introduce non-linearity. The updated feature matrix $\bm{H}^{(\ell+1)}$ is obtained by aggregating information from neighboring nodes up to $K$ hops away, using the weighted sum of the convolved features. In fact, each term $\hat{\bm{A}}_{k}\bm{H}^{(\ell)}$ represents the aggregation of information from nodes within $k$ hops of each node, weighted by the normalized $k$-adjacency matrix. This allows the model to capture multi-hop dependencies and incorporate information from distant nodes in the graph during the convolutional operation, thereby capturing both local and long-range dependencies in the graph structure. By disentangling the features under multi-hop aggregation, we ensure that the weights assigned to further neighborhoods are not biased due to their distance from the central node~\cite{Liu2020Disentangle}. This disentanglement process allows our model to effectively capture graph-wide joint relationships on human skeletons, leading to more accurate modeling of long-range dependencies and improving the performance of GCNs in capturing complex relationships between body joints.

The key advantage of using the $k$-adjacency matrix is that it allows for flexible modeling of long-range relationships between body joints. Also, one important characteristic of the $k$-adjacency matrices is that the relationships they represent have low correlations with each other. This means that the relationships captured by the $k$-adjacency matrices of different hops are distinct and provide unique information. As a result, the $k$-adjacency matrix structure enables the modeling of diverse and long-range relationships between body joints, even in cases where direct neighbors might not exist or are limited within the specified hop distance. By considering further neighborhoods, the $k$-adjacency matrix provides a broader view of the graph structure and captures more distant dependencies between nodes. It allows for modeling long-range relationships and capturing complex interactions among nodes that may not be apparent in the standard adjacency matrix with self-added loops.

By incorporating additional hops with larger values of $k$, the aggregation rule of the multi-hop GCN aggregates information in an additive manner, allowing for effective modeling of long-range dependencies, as the model can capture relationships between joints that are further apart. In terms of sparsity, the $k$-adjacency matrix is generally more sparse than the $k$-th power of the adjacency matrix, especially for large values of $k$, as shown in Figure~\ref{Fig:AdjacencyMatrices}. In general, as $k$ increases, the number of non-zero entries in the $k$-th power of the adjacency matrix tends to increase, potentially leading to a denser representation. Hence, the sparsity property of the $k$-adjacency matrix is beneficial as it reduces the computational complexity and memory requirements of the model, resulting in more efficient representations.

\begin{figure}[!htb]
\centering
\includegraphics[scale=.23]{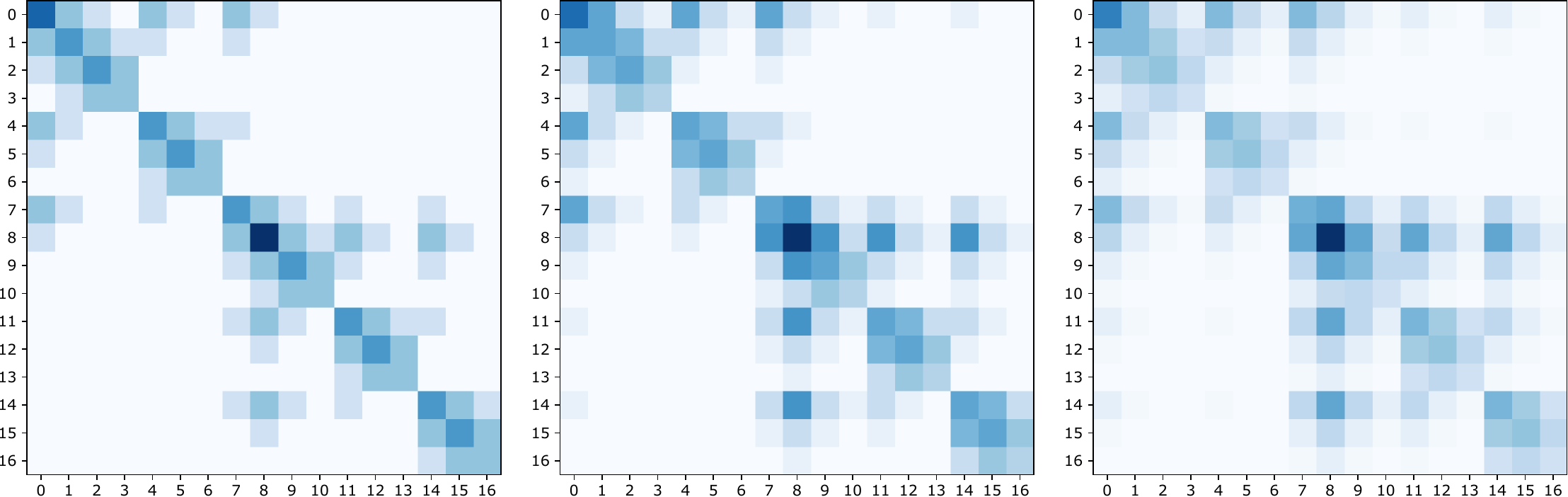} \\
\hspace{.2in}$\tilde{\bm{A}}^{2}$ \hspace{1in} $\tilde{\bm{A}}^{3}$ \hspace{.9in} $\tilde{\bm{A}}^{4}$ \hspace{.1in} \\[2ex]
\includegraphics[scale=.23]{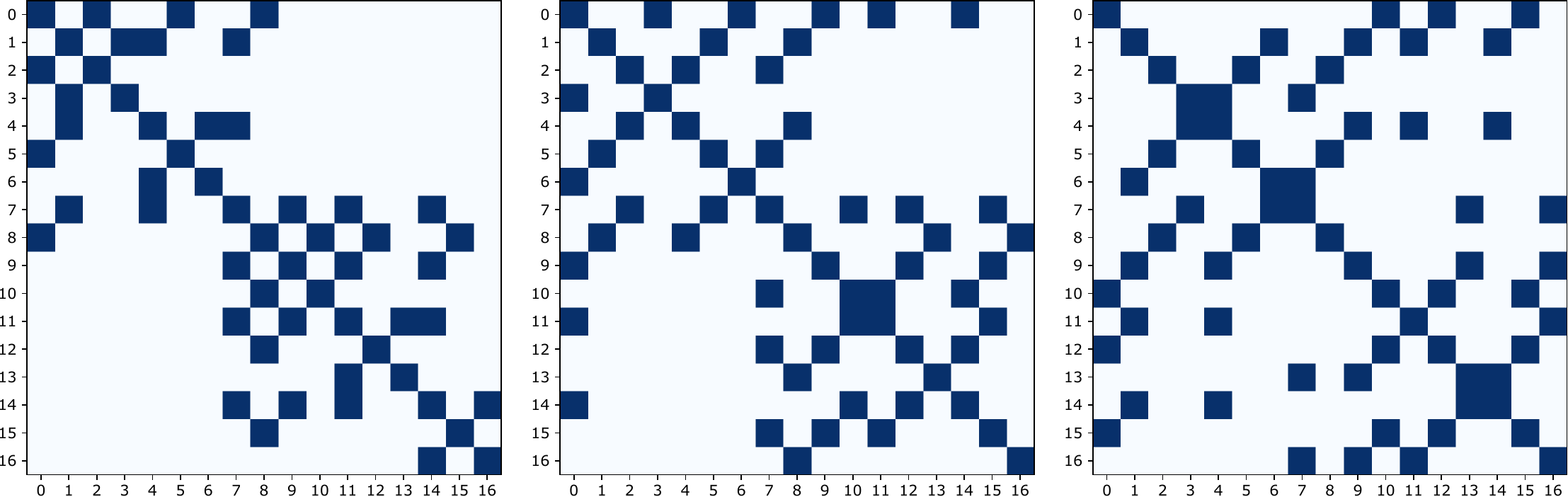} \\
\hspace{.2in}$\tilde{\bm{A}}_{2}$ \hspace{1in} $\tilde{\bm{A}}_{3}$ \hspace{.9in} $\tilde{\bm{A}}_{4}$ \hspace{.1in}
\caption{Comparing the sparsity of the $k$-th power of the adjacency matrix (top row) and the $k$-adjacency matrix (bottom row). As the value of $k$ increases, the $k$-th power representation tends to become denser, while the $k$-adjacency matrix maintains higher sparsity. The sparsity of the $k$-adjacency matrix makes it an efficient choice for capturing long-range dependencies in the multi-hop GCN with disentangled neighborhoods, reducing computational complexity and memory usage.}
\label{Fig:AdjacencyMatrices}
\end{figure}

\subsection{Multi-hop Graph Transformer Network}
The overall framework of our proposed MGT-Net is shown in Figure~\ref{Fig:NetworkArchitecture}. The network architecture is comprised of three main components: skeleton embedding, a graph attention block and a multi-hop graph convolutional block. The input is a sequence of 2D human poses, which are obtained using an off-the-shelf 2D detector [43], and the output is a 3D human pose. The multi-head self-attention mechanism enables our model to capture long-range dependencies and encode global context information, while multi-hop graph convolutions leverage the graph structure of body joints to exchange information between non-neighboring joints, allowing for the modeling of higher-order relationships.

\begin{figure*}[!ht]
\centering
\includegraphics[scale=.57]{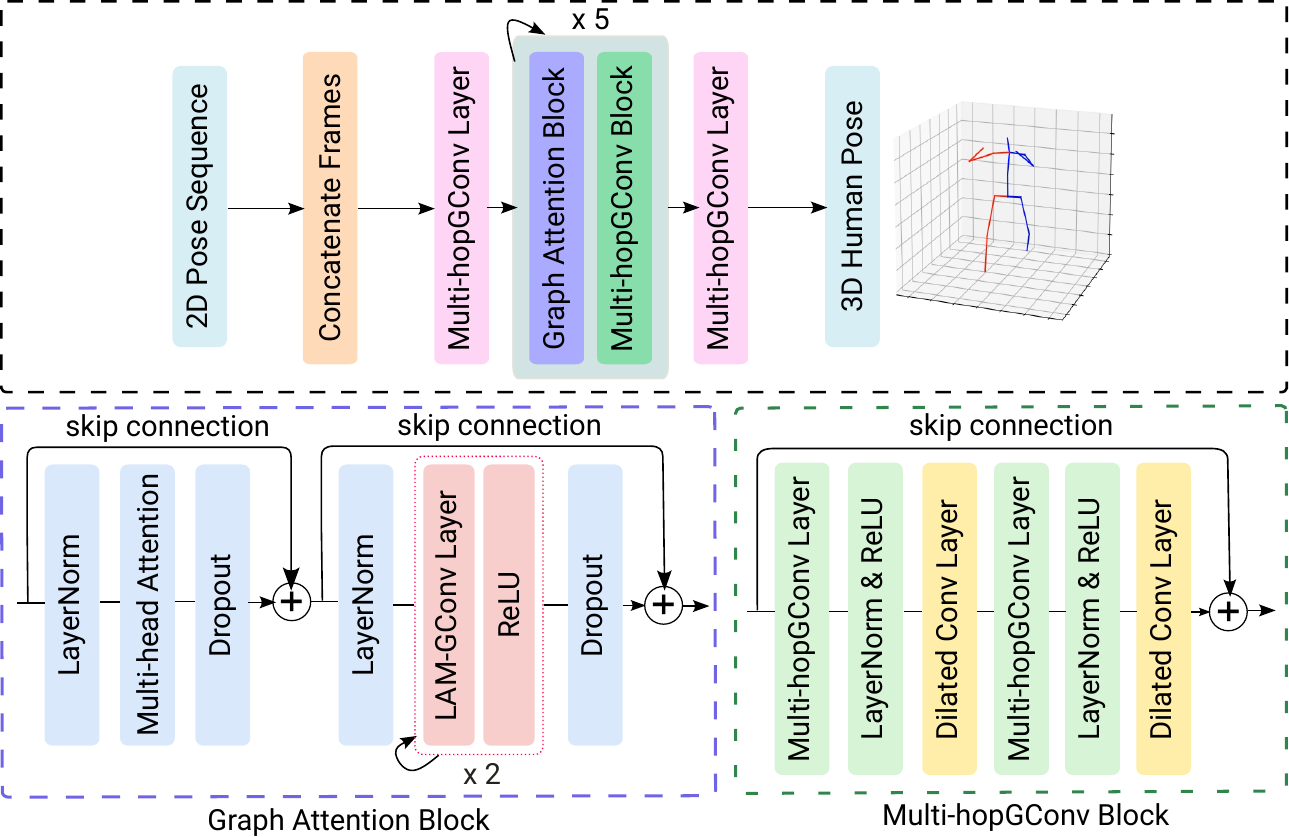}
\caption{Network architecture of the proposed MGT-Net for 3D human pose estimation. Our model takes a sequence of 2D pose coordinates as input and generates 3D pose predictions as output. The core building blocks of the network are a graph attention block and a multi-hop graph convolutional block, which are stacked together. We use a total of five layers for these stacks. In the graph attention block, the multi-head attention layer is followed by two consecutive graph convolutional layers with learnable adjacency matrix (LAM-GConv). The multi-hop graph convolutional block is composed of two subblocks, each of which comprises a multi-hopGConv layer, followed by a dilated convolutional layer.}
\label{Fig:NetworkArchitecture}
\end{figure*}

\subsubsection{Skeleton Embedding}
In order to incorporate temporal information into our model, we take a 2D pose sequence as input. Specifically, given a 2D pose sequence $\bm{S}\in \mathbb{R}^{N\times 2\times T}$ represented as a tensor, where $T$ denotes the number of frames and $N$ is the number of joints, we first reshape it into a matrix $\tilde{\bm{S}}\in \mathbb{R}^{N \times 2T}$ by concatenating the 2D coordinates of all frames along the time axis. This allows us to consider the 2D coordinates of all frames for each joint as a continuous sequence, effectively capturing the temporal dependencies and changes of the joints over time. Then, we pass it through a multi-hop graph convolution layer, resulting in an $N\times F$ embedding matrix $\bm{X}$ of joint attributes whose $i$-th row is an $F$-dimensional feature vector associated to the $i$-th joint of the skeleton graph. This skeleton embedding matrix serves as a fundamental representation that incorporates both spatial and temporal information from the 2D pose sequence. It forms the basis for subsequent components of our network architecture, facilitating the integration of both spatial and temporal context into the model.

\subsubsection{Graph Attention Block}
The graph attention block consists of stacked layers of multi-head self-attention and graph convolution, leveraging a learnable adjacency matrix. This block allows our model to capture both global and local dependencies within the graph structure. The multi-head self-attention mechanism enables the model to weigh the importance of different nodes, while the graph convolutional layers help propagate information across neighboring nodes.

\medskip\noindent\textbf{Multi-head Self-Attention.}\quad At their core, Transformers~\cite{Vaswani2017Tranformers} rely on self-attention, which allows the model to weigh the importance of different tokens within a given input sequence when making predictions. Self-attention operates on the embedding matrix $\bm{X}\in\mathbb{R}^{N\times F}$ consisting of $N$ feature vectors, each of which has an embedding dimension $F$. The input matrix $\bm{X}$ is first linearly projected into a query matrix $\bm{Q}$, a key matrix $\bm{K}$ and a value matrix $\bm{V}$ as follows:
\begin{equation}
\bm{Q}=\bm{X}\bm{W}_{q},\quad \bm{K}=\bm{X}\bm{W}_{k}, \bm{V}=\bm{X}\bm{W}_{v},
\end{equation}
where $\bm{W}_{q}\in\mathbb{R}^{F\times d_k}$, $\bm{W}_{k}\in\mathbb{R}^{F\times d_k}$ and $\bm{W}_{v}\in\mathbb{R}^{F\times d_k}$ are learnable weight matrices, and $d_k$ is the projection dimension. Then, the self-attention (SA) output is the weighted sum of the value vectors for each token
\begin{equation}
\text{SA}(\bm{X})=\text{softmax}\left(\frac{\bm{Q}\bm{K}^{\T}}{\sqrt{d_k}}\right)\bm{V}\in\mathbb{R}^{N\times d_{k}},
\end{equation}
where the weights are attention scores computed as the dot product between the query and key vectors, scaled by the square root of the projection dimension, and followed by softmax applied row-wise. These attention weights determine the importance of each value vector, enabling the model to capture contextual dependencies. Scaling the dot product between the query and key vectors by $\sqrt{d_k}$ helps stabilize the attention weights and prevents them from becoming too large, which can lead to numerical instability during training. This ensures that the attention weights are more balanced and can better capture the relationships between different feature vectors.

In practice, Transformers employ multi-head attention to capture different types of relationships and patterns in the input sequence. This involves performing multiple self-attention operations in parallel, each with its own set of weight matrices, allowing the model to capture different types of dependencies and attending to different parts of the input sequence. For simplicity, we assume $d_k=F/h$, where $h$ is the number of attention heads (i.e., self-attention operations). The outputs of $h$ heads are concatenated and linearly transformed using an $F\times F$ learnable weight matrix $\bm{W}_{o}$ to obtain the output of the multi-head self-attention (MSA) as follows:
\begin{equation}
\text{MSA}(\bm{X})=\text{Concat}(\bm{Y}_1,\dots,\bm{Y}_h)\bm{W}_{o},
\end{equation}
where $\bm{Y}_i=\text{SA}_{i}(\bm{X})\in\mathbb{R}^{N\times\frac{F}{h}}$ is the output of the $i$-th attention head. Hence, the final output of concatenated attention heads is $\bm{Y}=\text{MSA}(\bm{X})\in\mathbb{R}^{N\times F}$.

Taking inspiration from the architectural design of Graformer~\cite{zhao2022graformer}, our graph attention block is composed of a multi-head attention layer and two multi-hop graph convolutional layers. Each convolutional layer is equipped with a learnable adjacency matrix (LAM-GConv), which enables the model to adaptively learn the relationships between joints, making the graph convolution operation more flexible and effective in capturing local interactions. The combination of the multi-head attention layer and the LAM-GConv layers in our graph attention block allows for the integration of both global and local information, enhancing the model's ability to exploit the relationships among joints and capture important features for 3D human pose estimation. The output $\bm{Y}$ of the multi-head self-attention is passed through a LAM-GConv layer, which applies graph convolution to aggregate information from neighboring joints based on the learnable adjacency matrix. This step helps refine and update the joint features by considering their relationships with adjacent joints. We also apply a layer normalization layer (LayerNorm) to normalize the features across the joint dimension and a dropout layer to prevent overfitting and improve generalization. Moreover, a skip connection is employed to facilitate information flow, enabling the model to preserve important features from earlier stages of processing and pass them directly to subsequent layers.

\subsubsection{Multi-hop Graph Convolutional Block}
The multi-hop graph convolutional block combines multi-hop convolutional and dilated convolutional layers. This block facilitates the modeling of long-range dependencies and captures spatial relationships across different hops in the graph. A key benefit of using the dilated convolutional layer is that it can capture contextual information from a wider context thanks, in large part, to its larger receptive field. This is particularly useful in tasks where long-range dependencies and global patterns are important, such as 3D human pose estimation.

\medskip\noindent\textbf{Multi-hop Graph Convolution.}\quad The multi-hop graph convolutional layer exchanges information between non-neighboring nodes, enabling the model to capture broader context and larger receptive fields, incorporating the influence of distant joints on the pose estimation. This is particularly important for understanding the global structure and relationships among different parts of the graph, even when they are not directly connected in the graph representation. Also, the ability to aggregate information from different hops allows the model to effectively learn and represent the hierarchical structure of the pose, capturing both local details and global patterns. Moreover, by focusing on distinct and non-redundant neighborhoods, the multi-hop graph convolutional layer with disentangled neighborhoods can better discern meaningful information and discard irrelevant signals.

\medskip\noindent\textbf{Dilated Convolution.}\quad In traditional convolutional networks, the receptive field is determined by the kernel size, which directly affects the size of the local neighborhood that each convolutional operation considers. In contrast, dilated convolutional networks utilize dilated kernels to increase the receptive field of the convolutional layer~\cite{Yu2016Multiscale}. Specifically, given a 2D input $\bm{X}$ and a kernel filter $\bm{w}$ of size $(2m+1)\times (2m+1)$, the dilated convolution, denoted by $\ast_{d}$, is defined as
\begin{equation}
(\bm{X}\ast_{d}\bm{w})(i,j)=\sum_{r,s=-m}^{m}\bm{X}(i+d\times r, j+d\times s)\bm{w}(r,s),
\end{equation}
where $d$ is a dilation rate parameter, which controls the spacing between the kernel elements. By increasing the dilation rate, the kernel can effectively ``expand'' and cover a larger region of the input feature map. This expansion leads to an increased receptive field without the need for larger kernel sizes.

By applying a dilated convolutional layer after a multi-hop graph convolutional layer, we can leverage the strengths of both operations. The multi-hop graph convolutional layer captures local relationships and structural dependencies among nodes, while the dilated convolutional layer further enhances the receptive field by incorporating larger contextual information without increasing the number of parameters. This is crucial for 3D human pose estimation, as it allows the model to consider the relationships and dependencies between body joints across a broader region, providing a more comprehensive understanding of the pose. Moreover, dilated convolutions are effective at capturing long-range dependencies by considering information from distant nodes of the graph, and also at handling occlusion scenarios where body joints may be temporarily hidden or obscured.

\subsection{Model Training}
The parameters (i.e., weight matrices for different layers) of the proposed MGT-Net for 3D human pose estimation are learned by minimizing the following loss function
\begin{equation}
\mathcal{L} =\frac{1}{N}\biggl[(1-\alpha)\sum_{i=1}^{N}\Vert\bm{y}_{i}-\hat{\bm{y}}_{i}\Vert_{2}^{2} + \alpha\sum_{i=1}^{N}\Vert\bm{y}_{i}-\hat{\bm{y}}_{i}\Vert_{1}\biggr],
\end{equation}
which is a weighted sum of the mean squared and mean absolute errors between the 3D ground truth coordinates $\bm{y}_{i}$ and estimated 3D
joint coordinates $\hat{\bm{y}}_{i}$ over a training set consisting of $N$ human poses. These estimated 3D poses are generated by the last multi-hop graph convolutional layer of MGT-Net, as illustrated in Figure~\ref{Fig:NetworkArchitecture}. The proposed loss function draws inspiration from the penalty function used in the elastic net regression model~\cite{HuiZou:05}, which is a weighted combination of lasso and ridge regularization. For the mean squared error, the squared differences between the predicted and ground truth coordinates are averaged, meaning that larger errors have a greater impact on the overall score. In other words, the mean squared error is more sensitive to outliers and penalizes larger errors more heavily than the mean absolute error, which is more robust to outliers and treats all errors equally. The weighting factor $\alpha$ balances the contribution of each loss term. When $\alpha=0$, our loss function reduces to the mean squared error (i.e., ridge regression) and when $\alpha=1$, it reduces to the mean absolute error (i.e., lasso regression). The value of the weighting factor is determined by performing a grid search over a range of possible values for $\alpha$. In our experiments, the best performance on the validation set is achieved with $\alpha=0.01$.

\section{Experiments}
In this section, we present a comprehensive evaluation of the proposed method by comparing it with competing baselines.

\subsection{Experimental Setup}
\noindent\textbf{Datasets.}\quad We conduct a comprehensive evaluation of our model on two standard benchmark datasets for 3D human pose estimation: Human3.6M~\cite{ionescu2013human3} and MPI-INF-3DHP~\cite{mehta2017monocular}. These datasets provide robust evaluation scenarios for assessing the model performance.
	
\medskip\noindent\textbf{Evaluation Protocols and Metrics.}\quad For Human3.6M, we adopt Protocol \#1 and Protocol \#2 for training and testing~\cite{martinez2017simple}. Protocol \#1 utilizes the mean per-joint position error (MPJPE) metric, whereas Protocol \#2 uses the Procrustes-aligned mean per-joint position error (PA-MPJPE) metric. Both protocols employ five subjects (S1, S5, S6, S7, and S8) for training and two subjects (S9 and S11) for testing. For MPI-INF-3DHP, we use the Percentage of Correct Keypoint (PCK) and the Area Under Curve (AUC)~\cite{pavlakos2018ordinal} as evaluation metrics. For the 2D pose detection, we employ the cascaded pyramid network (CPN)~\cite{chen2018cascaded} on Human3.6M, while for MPI-INF-3DHP, we utilize the ground truth 2D pose as input.

\medskip\noindent\textbf{Baseline Methods.}\quad We evaluate the performance of our MGT-Net against various state-of-the-art methods, including SemGCN~\cite{zhao2019semantic}, M\"{o}biusGCN~\cite{azizi20223d}, GroupGCN~\cite{zhang2022group}, MM-GCN~\cite{multihop2022}, MGCN~\cite{zou2021modulated}, GraFormer~\cite{zhao2022graformer}, VideoPose3D~\cite{pavllo20193d}, ST-GCN~\cite{YujunCai:19}, SRNet~\cite{zeng2020srnet}, Attention3D~\cite{liu2020attention}, Anatomy3D~\cite{chen2021anatomy}, GAST-Net~\cite{liu2021graph}, Skeletal-GNN~\cite{zeng2021learning}, HTNet~\cite{cai2023htnet}, GS-Net~\cite{Zaedul2023GSNet}, and PoseFormerV2~\cite{Zhao2023PoseFormerV2}.

\medskip\noindent\textbf{Implementation Details.}\quad We train our model using the AMSGrad optimizer for 30 epochs with an initial learning rate of 0.005, and a decay factor of 0.90 per 4 epochs. We set the batch size to 128, the number of layers $L=5$, and adopt $h=4$ heads for self-attention, the middle feature dimension $F=256$, the weighting factor $\alpha=0.01$, and the total number of input frames $T=243$.

\subsection{Results and Analysis}
\noindent\textbf{Quantitative Results on Human3.6M.}\quad We report the performance comparison results for all 15 actions, along with the average performance, in Table~\ref{Tab:Result1}. Our MGT-Net outperforms several state-of-the-art methods when using the detected 2D pose as input. As can be seen, our method achieves on average 44.1mm and 36.2mm in terms of MPJPE and PA-MPJPE, respectively, outperforming all the baselines. These findings demonstrate the model's competitiveness, which is largely attributed to the fact that the MGT-Net can better exploit joint connections through the multi-hop graph propagation rule and learn not only the global information from all the nodes, but also the explicit adjacency structure of nodes. Under Protocol \#1, Table 1 reveals that our MGT-Net model performs better than GAST-Net~\cite{liu2021graph} in 11 out of 15 actions, yielding 0.8mm error reduction on average, improving upon this best performing temporal baseline by a relative improvement of 1.78\%, while maintaining a fairly small number of learnable parameters. In comparison to MGCN~\cite{zou2021modulated}, our method consistently achieves superior performance across 14 out of 15 actions, demonstrating an average error reduction of 5.3mm. This improvement represents a significant relative enhancement of 10.73\% over the strong MGCN baseline.

Under Protocol \#2, Table~\ref{Tab:Result1} shows that our model, on average, reduces the error by 0.82\% compared to Anatomy3D~\cite{chen2021anatomy}, which is the best spatio-temporal baseline, and achieves better results in 10 out of 15 actions. Also, our method outperforms GroupGCN in all 15 actions, yielding a relative error reduction of 9.73\% in terms of PA-MPJPE. In addition, PoseFormerV2 achieves the highest average PA-MPJPE score, while our model achieves the second-best performance.

\begin{table*}[!htb]
\caption{Performance comparison of our model and baseline methods on Human3.6M under Protocol \#1 and Protocol \#2 using the detected 2D pose as input. MPJPE and PA-MPJPE errors are in millimeters. The average errors are reported in the last column. Boldface numbers indicate the best performance, whereas the underlined numbers indicate the second best performance. $T$ denotes the number of input frames used in each spatio-temporal method.}
\small
\setlength{\tabcolsep}{1.8pt}
\medskip
\centering
\begin{tabular}{l*{17}{c}}
\toprule[1.2pt]
& \multicolumn{15}{c}{Action}\\
\cmidrule(lr){2-16}
\textbf{Protocol \#1} & Dire. & Disc. &  Eat & Greet & Phone & Photo &  Pose & Purch. & Sit & SitD. & Smoke & Wait & WalkD. & Walk & WalkT. & Avg.\\
\midrule[.8pt]
SemGCN~\cite{zhao2019semantic} & 47.3& 60.7& 51.4 &60.5& 61.1& 49.9& 47.3 & 68.1 & 86.2 & 55.0 & 67.8 & 61.0 & 42.1 & 60.6 & 45.3 & 57.6\\
M\"{o}biusGCN~\cite{azizi20223d} & 46.7 & 60.7 & 47.3 & 50.7 & 64.1 & 61.5 & 46.2 & 45.3 & 67.1 & 80.4 & 54.6 & 51.4 & 55.4 & 43.2 & 48.6 & 52.1 \\
GroupGCN~\cite{zhang2022group} & 45.0 & 50.9 & 49.0 & 49.8 & 52.2 & 60.9 & 49.1 & 46.8 & 61.2 & 70.2 & 51.8 & 48.6 & 54.6 & 39.6 & 41.2 & 51.6 \\
MM-GCN~\cite{multihop2022} & 46.8 & 51.4 & 46.7 & 51.4 & 52.5 & 59.7 & 50.4 & 48.1 & 58.0 & 67.7 & 51.5 & 48.6 & 54.9 & 40.5 & 42.2 & 51.7 \\
MGCN~\cite{zou2021modulated} & 45.4 & 49.2 & 45.7 & 49.4 & 50.4 &  58.2 & 47.9 & 46.0 & 57.5 & 63.0 & 49.7 & 46.6 & 52.2 & 38.9 & 40.8 & 49.4\\
GraFormer~\cite{zhao2022graformer} & 45.2 & 50.8 & 48.0 & 50.0 & 54.9 & 65.0 & 48.2 & 47.1 & 60.2 & 70.0 & 51.6 & 48.7 & 54.1 & 39.7 & 43.1 & 51.8 \\
GS-Net~\cite{Zaedul2023GSNet} & 41.1 & 46.6 & 43.0 & 48.0 & 48.6 & 52.4 & 44.6 & 41.9 & 54.5 & 65.9 & 46.2 & 46.1 & 48.2 & 38.6 & 40.9 & 47.1 \\
Hossain \textit{et al.}~\cite{hossain2018exploiting} ($T=5$) & 44.2 & 46.7 & 52.3 & 49.3 & 59.9 & 59.4 & 47.5 & 46.2 & 59.9 & 65.6 & 55.8 & 50.4 & 52.3 & 43.5 & 45.1 & 51.9 \\
Skeletal-GNN~\cite{zeng2021learning} ($T=9$)  & - & - & - & - & - & - & - & - & - & - & - & - & - & - & - & 45.7 \\
ST-GCN~\cite{YujunCai:19} ($T=7$) & 44.6 & 47.4 & 45.6 & 48.8 & 50.8 & 59.0 & 47.2 & 43.9 & 57.9 & 61.9 & 49.7 & 46.6 & 51.3 & 37.1 & 39.4 & 48.8\\
HTNet~\cite{cai2023htnet} ($T=9$) & - & - & - & - & - & - & - & - & - & - & - & - & - & - & - & 47.2\\
Lin \textit{et al.}~\cite{lin2019trajectory} ($T=50$) & 42.5 & \underline{44.8} & 42.6 & 44.2 & 48.5 & 57.1 & 42.6 & 41.4 & 56.5 & 64.5 & 47.4 & \underline{43.0} & 48.1 & 33.0 & 35.1  & 46.6\\
VideoPose3D~\cite{pavllo20193d} ($T=243$) & 45.2 & 46.7 & 43.3 & 45.6 & 48.1 & 55.1 & 44.6 & 44.3 & 57.3 & 65.8 & 47.1 & 44.0 & 49.0 & 32.8 & 33.9  & 46.8\\
Attention3D~\cite{liu2020attention} ($T=243$) & 41.8 & \underline{44.8} & 41.1 & 44.9 & 47.4 & 54.1 & 43.4 & 42.2 & 56.2 & 63.6 & \underline{45.3} & 43.5 & 45.3 & \underline{31.3} & \underline{32.2} & 45.1 \\
SRNet~\cite{zeng2020srnet} ($T=243$) & 46.6 & 47.1 & 43.9 & \textbf{41.6} & \underline{45.8} & \underline{49.6} & 46.5 & \textbf{40.0} & 53.4 & 61.1 & 46.1 & \textbf{42.6} & \textbf{43.1} & 31.5 & 32.6 & \underline{44.8} \\
GAST-Net~\cite{liu2021graph} ($T=243$) & 43.3 & 46.1 & \underline{40.9} & 44.6 & 46.6 & 54.0 & 44.1 & 42.9 & 55.3 & \textbf{57.9} & 45.8 & 43.4 & 47.3 & \textbf{30.4} & \textbf{30.3}  & 44.9\\
Anatomy3D~\cite{chen2021anatomy} ($T=243$) & 42.5 & 45.4 & 42.3 & 45.2 & 49.1 & 56.1 & 43.8 & 44.9 & 56.3 & 64.3 & 47.9 & 43.6 & 48.1 & 34.3 & 35.2 & 46.6\\
PoseFormerV2~\cite{Zhao2023PoseFormerV2} ($T=243$) & - & - & - & - & - & - & - & - & - & - & - & - & - & - & -  & 45.2\\
\midrule[.8pt]

Ours ($T=243$)  & \textbf{38.7} & \textbf{43.9} & 42.3 & 43.8 & \textbf{44.8} & \textbf{48.1} & \underline{42.4} & \underline{41.2} & \textbf{52.6} & 63.8 & \textbf{43.5} & 42.7 & \underline{44.7} & 34.1 & 34.5 & \textbf{44.1} \\

\midrule[1.5pt]
\textbf{Protocol \#2} & Dire. & Disc. &  Eat & Greet & Phone & Photo &  Pose & Purch. & Sit & SitD. & Smoke & Wait & WalkD. & Walk & WalkT. & Avg.\\
\midrule[.8pt]
Li \textit{et al.}~\cite{ChenLiLee:2020} & 38.5 & 41.7 & 39.6 & 45.2 & 45.8 & 46.5 & 37.8 & 42.7 & 52.4 & 62.9 & 45.3 & 40.9 & 45.3 & 38.6 & 38.4 & 44.3\\
High-order GCN~\cite{zou2020high} & 38.6 &42.8& 41.8 &43.4 &44.6& 52.9& 37.5& 38.6 & 53.3 & 60.0 & 44.4 & 40.9 & 46.9 & 32.2 &37.9 & 43.7\\
HOIF-Net~\cite{quan2021higher} & 36.9 & 42.1&40.3 &42.1 &43.7 &52.7&37.9 &37.7 &51.5 &60.3  &43.9&39.4 & 45.4 & 31.9 & 37.8 & 42.9 \\
MM-GCN~\cite{multihop2022} & 35.7 & 39.6 & 37.3 & 41.4 & 40.0 & 44.9 & 37.6 & 36.1 & 46.5 & 54.1 & 40.9 & 36.4 & 42.8 & 31.7 & 34.7 & 40.3 \\
GroupGCN~\cite{zhang2022group} & 35.3 & 39.3 & 38.4 & 40.8 & 41.4 & 45.7 & 36.9 & 35.1 & 48.9 & 55.2 & 41.2 & 36.3 & 42.6 & 30.9 & 33.7 & 40.1 \\
MM-GCN~\cite{multihop2022} & 35.7 & 39.6 & 37.3 & 41.4 & 40.0 & 44.9 & 37.6 & 36.1 & 46.5 & 54.1 & 40.9 & 36.4 & 42.8 & 31.7 & 34.7 & 40.3 \\
MGCN~\cite{zou2021modulated} & 35.7 & 38.6 & 36.3 & 40.5 & 39.2 & 44.5 & 37.0 & 35.4 & 46.4 &  51.2 & 40.5 & 35.6 & 41.7 & 30.7 & 33.9 & 39.1\\
GS-Net~\cite{Zaedul2023GSNet} & 34.5 & 38.4 & 35.0 & 40.9 & 38.9 & 42.4 & 35.9 & 33.9 & 44.2 & 55.9 & 38.2 & 36.7 & 40.6 & 30.4 & 33.8 & 38.7 \\
Hossain \textit{et al.}~\cite{hossain2018exploiting} ($T=5$) & 36.9 & 37.9 & 42.8 & 40.3 & 46.8 & 46.7 & 37.7 & 36.5 & 48.9 & 52.6 & 45.6 & 39.6 & 43.5 & 35.2 & 38.5 & 42.0 \\
ST-GCN~\cite{YujunCai:19} ($T=7$) & 35.7 & 37.8 & 36.9 & 40.7 & 39.6 & 45.2 & 37.4 & 34.5 & 46.9 & \underline{50.1} & 40.5 & 36.1 & 41.0 & 29.6 & 33.2 & 39.0\\
Lin \textit{et al.}~\cite{lin2019trajectory} ($T=50$) & \textbf{32.5} & \textbf{35.3} & \underline{34.3} & \textbf{36.2} & 37.8 & 43.0 & \textbf{33.0} & \textbf{32.2} & 45.7 & 51.8 & 38.4 & \textbf{32.8} & \underline{37.5} & \underline{25.8} & 28.9  & 36.8\\
VideoPose3D~\cite{pavllo20193d} ($T=243$) & 34.1 & 36.1 & 34.4 & 37.2 & \underline{36.4} & \underline{42.2} & 34.4 & 33.6 & 45.0 & 52.5 & \underline{37.4} & 33.8 & 37.8 & \textbf{25.6} & \underline{27.3} & 36.5\\
Anatomy3D~\cite{chen2021anatomy} ($T=243$) & 33.6 & \underline{36.0} & 34.4 & \underline{36.6} & 37.5 & 42.6 & \underline{33.5} & 33.8 & \underline{44.4} & 51.0 & 38.3 & \underline{33.6} & 37.7 & 26.7 & 28.2 & 36.5\\
GAST-Net~\cite{liu2021graph} ($T=243$) & 34.9 & 37.5 & 34.9 & 38.3 & 37.4 & 44.0 & 34.4 & 34.6 & 45.1 & \textbf{48.0} & 49.3 & 34.8 & 37.7 & 26.2 & \textbf{27.1}  & 36.9\\
PoseFormerV2~\cite{Zhao2023PoseFormerV2} ($T=243$) & - & - & - & - & - & - & - & - & - & - & - & - & - & - & -  & \textbf{35.6}\\
\midrule[.8pt]
Ours ($T=243$)  & \underline{33.0} & 36.1 & \textbf{34.1} & 37.4 & \textbf{36.2} & \textbf{40.4} & 33.6 & \underline{32.4} & \textbf{44.1} & 54.4 & \textbf{36.5} & 34.5 & \textbf{36.2} & 26.4 & 27.4 & \underline{36.2} \\		
\bottomrule[1.2pt]
\end{tabular}
\label{Tab:Result1}
\end{table*}

\medskip\noindent\textbf{Cross-Dataset Results on MPI-INF-3DHP.}\quad In Table~\ref{Tab:mpi_3dhp_inf}, we report the quantitative comparison results of the MGT-Net using a single frame against strong baselines on the MPI-INF-3DHP dataset. We train our model on the Human3.6M dataset and evaluate its performance on the MPI-INF-3DHP dataset to test its generalization ability across different datasets. The results demonstrate that our approach consistently outperforms the baseline methods in both indoor and outdoor scenes, achieving the highest PCK and AUC scores. In comparison to the best performing baseline, our model exhibits relative improvements of 4.26\% and 7.75\% in terms of the PCK and AUC metrics, respectively. Despite being trained solely on indoor scenes from the Human3.6M dataset, our model delivers satisfactory results when applied to outdoor settings. This demonstrates the robust generalization capability of our approach, extending its performance to unseen scenarios and datasets. In addition, the improvements in both PCK and AUC metrics demonstrate that our proposed model excels in both joint localization accuracy and overall pose estimation quality.

\begin{table}[!htb]
\caption{Performance comparison of our model without pose refinement and baseline methods on the MPI-INF-3DHP dataset using PCK and AUC as evaluation metrics.}
\setlength{\tabcolsep}{3pt}
\medskip
\centering
\begin{tabular}{lcc}
\toprule[1pt]
Method & PCK ($\uparrow$) & AUC ($\uparrow$)\\
\midrule[.8pt]
Chen \textit{et al.}~\cite{li2019generating} & 67.9 & - \\
Yang \textit{et al.}~\cite{yang20183d} & 69.0 & 32.0 \\
Pavlakos \textit{et al.}~\cite{pavlakos2018ordinal}  & 71.9 & 35.3 \\
Habibie \textit{et al.}~\cite{Habibie:19}  & 70.4 & 36.0 \\
HOIF-Net~\cite{quan2021higher} & 72.8 &36.5 \\
SRNet~\cite{zeng2020srnet} & 77.6 & 43.8\\
GraphSH~\cite{xu2021graph} & 80.1 & 45.8 \\
GroupGCN~\cite{zhang2022group} & 81.1 & 49.9\\
MM-GCN~\cite{multihop2022} & 81.6 & 50.3\\
Skeletal-GNN~\cite{zeng2021learning} & 82.1 & 46.2\\
GS-Net~\cite{Zaedul2023GSNet} & \underline{84.3} & \underline{53.0}\\
\midrule[.8pt]
Ours & \textbf{85.6} & \textbf{54.2}\\
\bottomrule[1pt]
\end{tabular}
\label{Tab:mpi_3dhp_inf}
\end{table}

\medskip\noindent\textbf{Qualitative Results.}\quad In Figure~\ref{Fig:Qualitative}, we present qualitative results obtained by our proposed MGT-Net on the Human3.6M dataset. The visual comparison showcases the performance of our model in 2D-to-3D human pose estimation, particularly in challenging scenarios. It is evident from the results that our MGT-Net outperforms the baseline method MGCN~\cite{zou2021modulated} and achieves closer alignment with the ground truth poses. Our model demonstrates effectiveness in accurately estimating 3D human poses, even in challenging cases where occlusions are present. Notably, MGCN~\cite{zou2021modulated} struggles to accurately predict poses in difficult instances such as ``Photo'', ``Sitting'', ``SitDown'', and ``WalkDog'' due to the presence of occlusions. In contrast, our MGT-Net consistently delivers reliable pose predictions in these challenging scenarios.

\begin{figure*}[!htb]
\centering
\includegraphics[scale=.7]{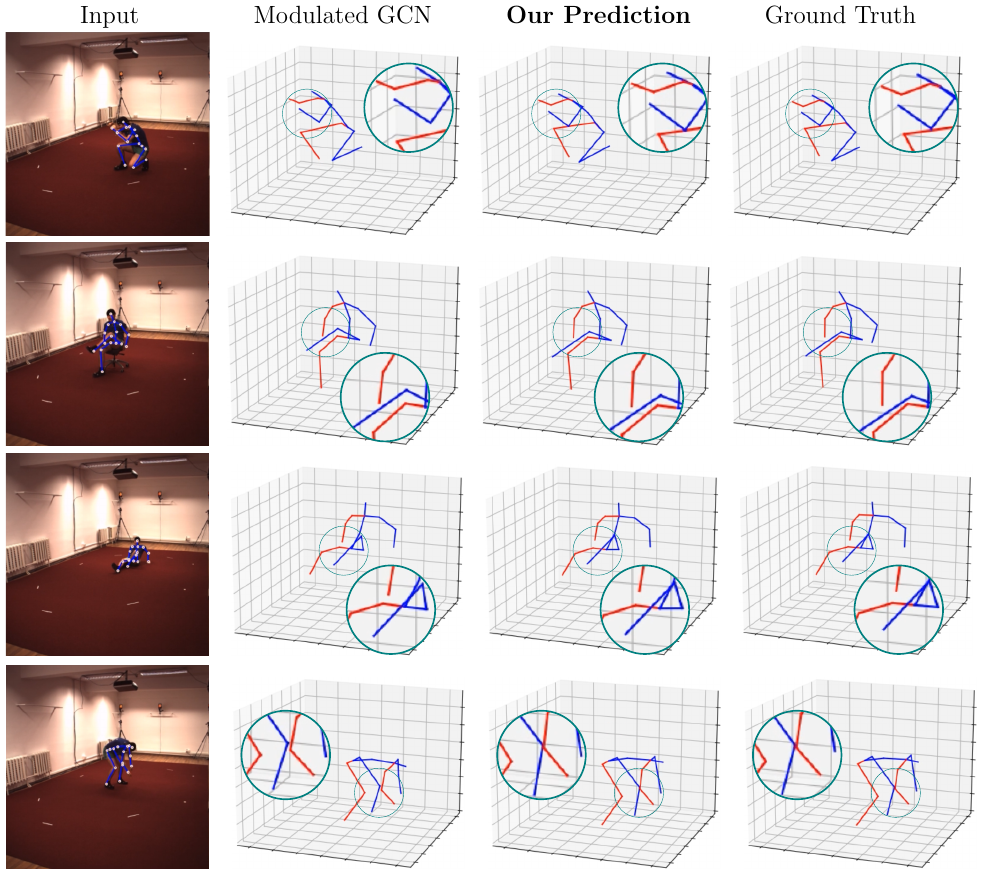}
\caption{Visual comparison between MGT-Net, MGCN and ground truth on the Human3.6M test set. Compared to MGCN, our model is able to produce better predictions.}
\label{Fig:Qualitative}
\end{figure*}

\medskip\noindent\textbf{Improvements on Hard Poses.}\quad Hard poses are characterized by high prediction errors and often exhibit certain inherent characteristics, such as depth ambiguity and self-occlusion~\cite{zeng2020srnet, zeng2021learning, zhao2019semantic}. For example, accurately estimating the 3D pose of a person sitting in a crossed-leg position can pose difficulties due to the intricate interactions among various body parts. Our method aims to address this challenge by learning to capture the complex relationships between the joints via the graph attention block and the multi-hop graph convolutional block with disentangled neighborhoods. As reported in Table~\ref{Tab:TemporalBaselines}, MGT-Net outperforms SRNet~\cite{zeng2020srnet} by reducing prediction errors in the actions ``Directions'', ``Eating'', ``Purchases'', ``Sitting'', and ``Sitting Down'' by 5.6mm, 1.0mm, 2.7mm, 4.2mm, and 1.0mm, respectively, which correspond to relative improvements of 16.10\%, 3.51\%, 8.85\%, 10.80\%, and 2.47\% under Protocol \#1. The average improvement on these hard poses is 8.35\%. Also, the visualization results shown in Figure~\ref{Fig:Qualitative} offer a comparative analysis with MGCN~\cite{zou2021modulated} on hard poses, demonstrating superior performance of our model. Hence, these quantitative and qualitative results demonstrate the effectiveness of our model in dealing with hard poses.

\begin{table*}[!htb]
\caption{Performance comparison of our model without pose refinement and spatio-temporal baseline methods on Human3.6M under Protocol \#1 using the ground truth 2D pose as input. $T$ denotes the number of input frames used in each spatio-temporal method.}
\setlength{\tabcolsep}{3pt}
\medskip
\small
\centering
\resizebox{1\textwidth}{!}{%
\begin{tabular}{l*{17}{c}}
\toprule
& \multicolumn{15}{c}{Action}\\
\cmidrule(lr){2-16}
\textbf{Protocol \#1} & Dire. & Disc. &  Eat & Greet & Phone & Photo &  Pose & Purch. & Sit & SitD. & Smoke & Wait & WalkD. & Walk & WalkT. & Avg.\\
\midrule[.8pt]
Hossain \textit{et al.}~\cite{hossain2018exploiting} ($T=5$) & 35.2 & 40.8 & 37.2 & 37.4 & 43.2 & 44.0 & 38.9 & 35.6 & 42.3 & 44.6 & 39.7 & 39.7 & 40.2 & 32.8 & 35.5 & 39.2 \\
Ray3D~\cite{Zhan_2022_CVPR} ($T=9$) & 31.2 & 35.7 & 31.4 & 33.6 & 35.0 & 37.5 & 37.2 & 30.9 & 42.5 & 41.3 & 34.6 & 36.5 & 32.0 & 27.7 & 28.9 & 34.4 \\
ST-GCN~\cite{YujunCai:19} ($T=7$) & 32.9 & 38.7 & 32.9 & 37.0 & 37.3 & 44.8 & 38.7 & 36.1 & 41.0 & 45.6 & 36.8 & 37.7 & 37.7 & 29.5 & 31.6 & 37.2\\
Lin \textit{et al.}~\cite{lin2019trajectory} ($T=50$) & - & - & - & - & - & - & - & - & - & - & - & - & - & - & -  & 32.8\\
PoseFormer~\cite{PoseFormer:2021} ($T=81$) & \underline{30.0} & 33.6 & 29.9 & 31.0 & \textbf{30.2} & \textbf{33.3} & 34.8 & 31.4 & 37.8 & \underline{38.6} & 31.7 & 31.5 & \textbf{29.0} & 23.3 & \textbf{21.1}  & \underline{31.3}\\
Attention3D~\cite{liu2020attention} ($T=243$) & 34.5 & 37.1 & 33.6 & 34.2 & 32.9 & 37.1 & 39.6 & 35.8 & 40.7 & 40.7 & 41.4 & 33.0 & 33.8 & 26.6 & 26.9 & 34.7 \\
SRNet~\cite{zeng2020srnet} ($T=243$) & 34.8 & \textbf{32.1} & 28.5 & \underline{30.7} & 31.4 & \underline{36.9} & 35.6 & 30.5 & 38.9 & 40.5 & 32.5 & 31.0 & 29.9 & \underline{22.5} & 24.5 & 32.0\\
GAST-Net~\cite{liu2021graph} ($T=243$) & 30.5 & 33.1 & \underline{27.6} & 31.0 & 31.8 & 37.0 & \textbf{33.2} & \underline{30.0} & \underline{35.7} & \textbf{37.7} & \underline{31.4} & \textbf{29.8} & 31.7 & 24.0 & 25.7  & 31.4\\
VideoPose3D~\cite{pavllo20193d} ($T=243$) & 35.2 & 40.2 & 32.7 & 35.7 & 38.2 & 45.5 & 40.6 & 36.1 & 48.8 & 47.3 & 37.8 & 39.7 & 38.7 & 27.8 & 29.5 & 37.8 \\
Anatomy3D~\cite{chen2021anatomy} ($T=243$) & - & - & - & - & - & - & - & - & - & - & - & - & - & - & - & 32.3\\
\midrule[.8pt]
Ours ($T=243$) & \textbf{29.2} & \underline{32.9} & \textbf{27.5} & \textbf{30.5} & \underline{30.8} & 37.1 & \underline{33.5} & \textbf{27.8} & \textbf{34.7} & 39.5 & \textbf{30.1} & \underline{30.3} & \underline{29.8} & \textbf{21.9} & \underline{23.1} & \textbf{30.6} \\
\bottomrule[1pt]
\end{tabular}
}
\label{Tab:TemporalBaselines}
\end{table*}

\subsection{Ablation Studies}
We conduct ablation studies on the Human3.6M dataset with the aim of analyzing the impact of different design choices in our network architecture. We use MPJPE and PA-MPJPE as evaluation metrics. Evaluation of our ablation experiments is performed without pose refinement, as we aim to assess the performance of our base model without additional refinement steps. To ensure a fair comparison, we train and test our model using 2D ground truth poses, eliminating any potential uncertainties introduced by 2D pose detectors. In our ablation experiments, we systematically vary specific parameters and/or components of our model and assess their influence on the overall performance. By conducting these controlled experiments, we can gain deeper insights into the importance of the key components of our model. Unless indicated otherwise, we set the number of input frames $T = 243$. This choice allows us to evaluate the model's performance with a significant temporal context.

\medskip\noindent\textbf{Impact of Input Sequence Length.}\quad  In Table~\ref{Tab:inputSequenceLength}, we report the MPJPE and PA-MPJPE results of our method using different input sequence lengths. We can observe that increasing the number of frames leads to better results. This is expected since temporal correlations help address challenges like depth ambiguity and self-occlusions, which are typically not easy to handle by single frame 3D pose estimation methods. It is also worth noting that the MPJPE and PA-MPJPE errors for $T=1$ are 35.0mm and 27.5mm, respectively. As $T$ increases, the errors decrease and the best results are obtained when $T=243$ for both protocols. We also report the number of model parameters for different input sequence lengths, and we can see that as $T$ increases, there is only a moderate increase in the number of learnable parameters. Specifically, when $T$ is set to 1 (i.e., single input frame), the total number of model parameters is 1.46M. However, as we progressively increase $T$ to 243, which represents a substantial number of input frames, the model parameters is only slightly increased to 1.65M, while achieving 12.57\% and 10.18\% relative error reductions in MPJPE and PA-MPJPE, respectively. This slight increase in model size is largely attributed to the fact that the number of frames mostly impacts the first Multi-hopGConv layer, which does not require a large number of learnable parameters. Hence, increasing the input sequence length yields improved results without significantly increasing the number of parameters.

\begin{table}[!htb]
\caption{Ablation study on the number of input frames ($T$).}
\medskip
\centering
\begin{tabular}{l*{7}{c}}
\toprule[1pt]
$T$ & 1  & 3 & 27 & 81 & 121 & 243\\
\midrule[.8pt]
MPJPE $(\downarrow)$ & 35.0 & 34.3 & 33.2 & 33.2 & 31.7 & \textbf{30.6} \\
PA-MPJPE $(\downarrow)$ & 27.5 & 27.7 & 26.9 & 26.8 & 25.2 & \textbf{24.7} \\
Params (M) & 1.46 & 1.46 & 1.48 & 1.52 & 1.55 & 1.65\\
\bottomrule[1pt]
\end{tabular}
\label{Tab:inputSequenceLength}
\end{table}

\medskip\noindent\textbf{Impact of Number of Hops.}\quad Table~\ref{Tab:differentHops} provides an analysis of the performance of our model with varying numbers of hops ($k$) in the multi-hop graph convolutional layer with disentangled neighborhoods. We observe that as the number of hops increases, the model's performance improves. Specifically, the 1-hop model outperforms the 0-hop model, resulting in a reduction of 0.2mm in both MPJPE and PA-MPJPE errors. Furthermore, the 2-hop model achieves even better performance than the 1-hop model, with a relative error reduction of 3.17\% in terms of MPJPE. This demonstrates the effectiveness of the multi-hop graph convolutional layer in capturing long-range interdependencies among body joints, enabling the model to better understand the spatial relationships and dependencies within the human pose. The results highlight the importance of incorporating multi-hop connections to enhance the model's ability to capture complex structural dependencies, leading to improved pose estimation accuracy.

\begin{table}[!htb]
\caption{Ablation study on the number of hops. The embedding dimension is set to $F=128$.}
\setlength{\tabcolsep}{4pt}
\medskip
\centering
\begin{tabular}{l*{4}{c}}
\toprule[1pt]
\# hops ($k$) & Params (M) & MPJPE $(\downarrow)$ & PA-MPJPE $(\downarrow)$\\
\midrule[.8pt]
$0$-hop & 1.19 & 31.8 & 25.1 \\
$1$-hop & 1.42 & 31.6 & 24.9 \\
$2$-hop & 1.65 & \textbf{30.6} & \textbf{24.7} \\
\bottomrule[1pt]
\end{tabular}
\label{Tab:differentHops}
\end{table}

\medskip\noindent\textbf{Impact of Dilated Convolutional Layer.} \quad We conducted an ablation experiment to analyze the impact of incorporating dilated convolutional layers (DCLs) after each Multi-hop GConv layer in the multi-hop graph convolutional block. The results, as shown in Table~\ref{Tab:DilatedConv}, demonstrate that the inclusion of DCLs significantly improves the performance of the model while maintaining computational efficiency. Notably, we observe a reduction of 1.5mm in MPJPE and 0.6mm in PA-MPJPE errors without a substantial increase in the number of learnable parameters. This improvement can be attributed, in part, to the ability of dilated convolutions to effectively increase the receptive field without adding more parameters or computational cost. By adjusting the dilation rate, the convolutional kernel can capture information from a wider region of the input, allowing the model to capture larger contextual information. This capability is particularly important in 3D pose estimation, where understanding the relationship between body joints across different scales and distances is crucial. By effectively capturing long-range dependencies and global context, dilated convolutions contribute to a more accurate estimation of 3D human poses, especially in scenarios involving complex poses.
\begin{table}[!htb]
\caption{Ablation study on dilated convolutional layer (DCL). The embedding dimension is set to $F=128$.}
\setlength{\tabcolsep}{3pt}
\medskip
\centering
\begin{tabular}{l*{4}{c}}
\toprule[1pt]
Method & Params (M) & MPJPE $(\downarrow)$ & PA-MPJPE $(\downarrow)$ \\
\midrule[.8pt]
w/o DCL & 1.48 & 32.1 & 25.3 \\
Ours & 1.65 & \textbf{30.6} & \textbf{24.7} \\
\bottomrule[1pt]
\end{tabular}
\label{Tab:DilatedConv}
\end{table}

\medskip\noindent\textbf{Impact of Graph Convolutional Layers.}\quad We performed a comparative analysis of high-order and multi-hop GCNs, and the results on the Human3.6M dataset are summarized in Table~\ref{Tab:MultiHopHigherOrder} using $K=2$ for graph convolutions. Our findings reveal that the proposed multi-hop GCN yields improved performance compared to its high-order counterpart under both evaluation protocols.

\begin{table}[!htb]
\caption{Performance comparison between high-order GCN and multi-hop GCN on Human3.6M under Protocols \#1 and \#2. The number of hops is set to $K=2$.}
\small
\setlength{\tabcolsep}{3pt}
\medskip
\centering
\begin{tabular}{l*{7}{c}}
\toprule[1pt]
Method & Params ($M$) & MPJPE $(\downarrow)$ & PA-MPJPE $(\downarrow)$ \\
\midrule[.8pt]
High-order GCN & 1.65 & 32.7 & 26.1 \\
Ours (Multi-hop GCN) & 1.65 & \textbf{30.6} & \textbf{24.7} \\
\bottomrule[1pt]
\end{tabular}
\label{Tab:MultiHopHigherOrder}
\end{table}

\subsection{Hyperparameter Sensitivity Analysis}
Our findings regarding the impact of various hyperparameters on model performance are summarized in Table~\ref{Tab:Hyperparameter}. From the results, it can be observed that a batch size of $B=128$ yields the best performance. The value of the hidden dimension $F$ has an impact on the model's performance in terms of of capturing patterns. Increasing $F$ from 96 to 128 leads to a decrease in both MPJPE and PA-MPJPE from 32.9mm and 26.0mm to 30.6mm and 24.7mm, respectively, along with a reasonable increase in the number of model parameters. However, further increasing $F$ to 256 results in a significant increase in the number of learnable parameters (from 1.65M to 5.78M) with a noticeable degradation in performance under both protocols. Among different values of the number of attention heads, $h=4$ produces the best results in terms of MPJPE (30.6mm) and PA-MPJPE (24.7mm) compared to $h=2$ and $h=8$. As for the number of layers, starting with $L=3$ and increasing it to $L=5$ yields the best results. Therefore, the best performance on Human3.6M with ground truth 2D pose as input is achieved using hyperparameter values of $L=3$, $F=128$, $B=128$, and $h=4$.

\begin{table}[!htb]
\caption{Ablation study on various configurations of our model:$L$ is the number of MGT-Net layers, $F$ is the hidden dimension of skeleton embedding, $B$ is the batch size, and $h$ is the number of attention heads.}
\setlength{\tabcolsep}{3.1pt}
\medskip
\centering
\begin{tabular}{lrrcccc}
\toprule[1pt]
$L$ & $F$ & $B$ & $h$ & Params (M) & MPJPE $(\downarrow)$ & PA-MPJPE $(\downarrow)$ \\
\midrule[.8pt]
5 & 96 & 128 & 4 & 1.02 & 32.9 & 26.0\\
5 & 128 & 128 & 4 & 1.65 & \textbf{30.6} & \textbf{24.7}\\
5 & 256 & 128 & 4 & 5.78 & 31.2 & 25.3\\
3 & 128 & 128 & 4 & 1.12 & 31.8 & 26.0\\
4 & 128 & 128 & 4 & 1.38 & 31.2 & 25.3\\
6 & 128 & 128 & 4 & 1.91 & 32.0 & 25.4\\
7 & 128 & 128 & 4 & 2.18 & 31.9 & 25.0\\
5 & 128 & 128 & 2 & 1.65 & 32.5 & 25.7\\
5 & 128 & 128 & 8 & 1.65 & 31.3 & 25.3\\
5 & 128 & 64 & 4 & 1.65 & 30.6 & 25.2\\
5 & 128 & 256 & 4 & 1.65 & 31.6 & 25.2\\
5 & 128 & 512 & 4 & 1.65 & 30.9 & 25.4\\
\bottomrule[1pt]
\end{tabular}
\label{Tab:Hyperparameter}
\end{table}

\subsection{Model Efficiency}
In Table~\ref{Tab:Runtime}, we present a comprehensive analysis of the computational efficiency and performance of our model compared to state-of-the-art baselines. We assess the trade-off between the model's computational cost and its performance in terms of the number of input frames ($T$), total number of parameters, estimated number of floating-point operations (FLOPs), and MPJPE. The evaluation is conducted on the Human3.6M dataset using both detected 2D poses and ground truth as inputs. Notably, our model achieves a balance between computational efficiency and accuracy, outperforming strong baselines while maintaining a relatively low computational cost. For the detected 2D poses, our model achieves an MPJPE of 44.1, demonstrating superior performance compared to the baselines. Furthermore, when provided with ground truth 2D poses as input, our model achieves a significantly reduced MPJPE of 30.6, showcasing the effectiveness of our approach.

\begin{table}[!htb]
\caption{Efficiency of our model in comparison with baselines in terms of the number of input frames ($T$), total number of parameters, FLOPs, and MPJPE. Evaluation is performed on Human3.6M using both the detected 2D poses and ground-truth (GT) as inputs.}
\setlength{\tabcolsep}{2.5pt}
\medskip
\centering
\begin{tabular}{lrrrc}
\toprule[1pt]
Method & $T$ &  Params (M)  & FLOPs (M) &  MPJPE $(\downarrow)$\\
\midrule[.8pt]
ST-GCN~\cite{YujunCai:19} & 7 & 5.18 & 469.81 & 48.8\\
PoseFormer~\cite{PoseFormer:2021} & 81 & 9.60 & 1358 & 44.3\\
CrossFormer~\cite{hassanin2022crossformer} & 27 & 9.93 & 515 & 46.5\\
Anatomy3D~\cite{chen2021anatomy} & 81 & 45.53 & 88.9 & 44.6\\
\midrule[.8pt]
Ours & 243 & 5.78 & 84.41 & \textbf{44.1} \\
Ours (GT) & 243 & 1.65 & 21.23 & \textbf{30.6} \\
\bottomrule[1pt]
\end{tabular}
\label{Tab:Runtime}
\end{table}

\subsection{Limitations}
Transformer-based methods, including our MGT-Net model, while promising for 3D human pose estimation, have two main limitations worth noting~\cite{Zhao2023PoseFormerV2}. One key limitation is the length of the input joint sequence, as these methods often apply self-attention across all frames of the input sequence, leading to significant computational overhead when the number of frames increases for improved estimation accuracy. Another limitation is the lack of robustness to noise inherent in the output of 2D joint detectors. These challenges can be alleviated by adopting a trajectory representation based on the discrete cosine transform (DCT). Specifically, the input skeleton sequence is transformed into the frequency domain using DCT~\cite{Mao2019DCT,Zhao2023PoseFormerV2} and then only a subset of low-frequency coefficients is retained. The rationale behind this approach is that by discarding high frequencies, DCT offers a more concise representation that adeptly captures the smoothness of human motion. To gain further insights into the constraints of our model, we also investigated instances where MGT-Net did not perform as expected, and the results are presented in Figure~\ref{Fig:FailureCases}, which illustrates failure cases of our model predictions for the ``Walking'' and ``Greeting'' actions from the Human3.6M dataset. Notice that our predictions do not align perfectly with the ground truth poses, particularly in situations characterized by self-occlusions.

\begin{figure}[!htb]
\centering
\setlength{\tabcolsep}{5pt}
\begin{tabular}{cc}
\includegraphics[width=3.5in]{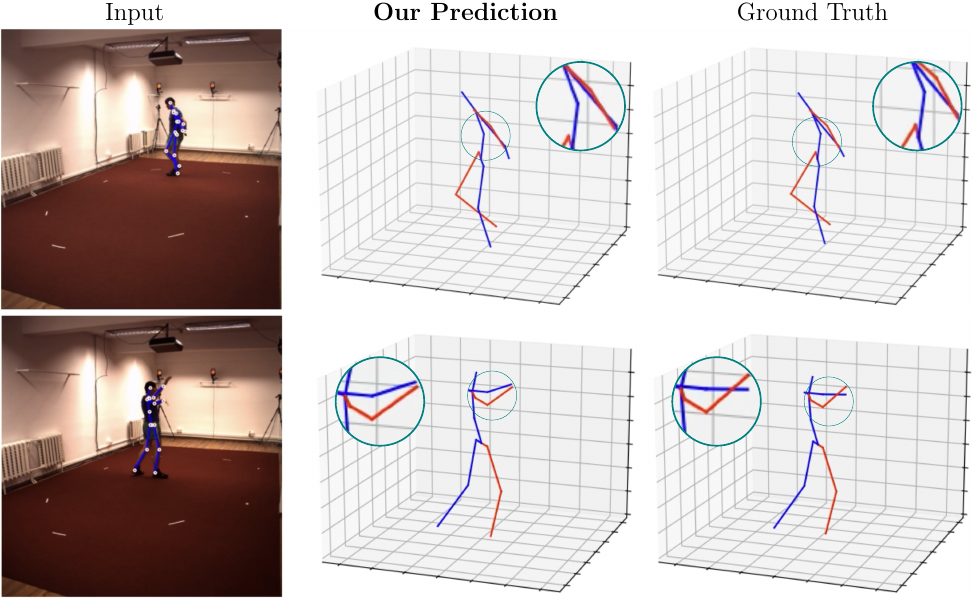}
\end{tabular}
\caption{Instances of the failure cases of our model on the ``Walking'' and ``Greeting'' actions from Human3.6M.}
\label{Fig:FailureCases}
\end{figure}

\section{Conclusion}
In this paper, we proposed MGT-Net, a spatio-temporal model for 3D human pose estimation. By leveraging multi-head self-attention, our model is able to focus on different aspects of the input data simultaneously, allowing it to capture intricate spatial relationships between body joints. On the other hand, the integration of multi-hop graph convolutions with disentangled neighborhoods enables our model to aggregate information from neighboring nodes at different hops, capturing both local and global contextual cues in the graph structure. To further enhance the model's understanding of spatial relationships and dependencies, we incorporated dilated graph convolutions into our network architecture. This integration extends the model's receptive field, allowing it to capture larger contextual information and better comprehend the relationships between body joints at varying scales and distances. Through extensive experiments and ablation studies, we demonstrated the effectiveness of our model compared to strong baselines. For future work, we aim to further enhance our approach through architectural improvements and explore its extension to 3D multi-person pose estimation.

\bigskip\noindent\textbf{Acknowledgments.}\quad This work was supported in part by the Discovery Grants program of Natural Sciences and Engineering Research Council of Canada.

\bibliographystyle{ieeetr}
\bibliography{references} 

\begin{thebibliography}{10}

\bibitem{Song2021Survey}
L.~Song, G.~Yu, J.~Yuan, and Z.~Liu, ``Human pose estimation and its
  application to action recognition: A survey,'' {\em Journal of Visual
  Communication and Image Representation}, vol.~76, 2021.

\bibitem{Luvizon2021Action}
D.~C. Luvizon, D.~Picardu, and H.~Tabia, ``Multi-task deep learning for
  real-time {3D} human pose estimation and action recognition,'' {\em IEEE
  Transactions on Pattern Analysis and Machine Intelligence}, vol.~43,
  pp.~2752--2764, 2021.

\bibitem{Zanfir2023Auto}
A.~Zanfir, M.~Zanfir, A.~Gorban, J.~Ji, Y.~Zhou, D.~Anguelov, and
  C.~Sminchisescu, ``{HUM3DIL}: Semi-supervised multi-modal {3D} human pose
  estimation for autonomous driving,'' in {\em Proc. Conference on Robot
  Learning}, 2023.

\bibitem{Ingwerse2032Sport}
C.~K. Ingwersen, C.~Mikkelstrup, J.~N. Jensen, M.~R. Hannemose, and A.~B. Dahl,
  ``{SportsPose} -- a dynamic {3D} sports pose dataset,'' in {\em Proc. IEEE
  Conference on Computer Vision and Pattern Recognition Workshops}, 2023.

\bibitem{Gu2019Therapy}
Y.~Gu, S.~Pandit, E.~Saraee, T.~Nordahl, T.~Ellis, and M.~Betke, ``Home-based
  physical therapy with an interactive computer vision system,'' in {\em Proc.
  IEEE Conference on Computer Vision and Pattern Recognition Workshops}, 2019.

\bibitem{zhou2016deep}
X.~Zhou, X.~Sun, W.~Zhang, S.~Liang, and Y.~Wei, ``Deep kinematic pose
  regression,'' in {\em Proc. European Conference on Computer Vision},
  pp.~186--201, 2016.

\bibitem{park20163d}
S.~Park, J.~Hwang, and N.~Kwak, ``{3D} human pose estimation using
  convolutional neural networks with {2D} pose information,'' in {\em Proc.
  European Conference on Computer Vision}, pp.~156--169, Springer, 2016.

\bibitem{sun2018integral}
X.~Sun, B.~Xiao, F.~Wei, S.~Liang, and Y.~Wei, ``Integral human pose
  regression,'' in {\em Proc. European Conference on Computer Vision},
  pp.~529--545, 2018.

\bibitem{pavlakos2017coarse}
G.~Pavlakos, X.~Zhou, K.~G. Derpanis, and K.~Daniilidis, ``Coarse-to-fine
  volumetric prediction for single-image {3D} human pose,'' in {\em Proc. IEEE
  Conference on Computer Vision and Pattern Recognition}, pp.~7025--7034, 2017.

\bibitem{sun2017compositional}
X.~Sun, J.~Shang, S.~Liang, and Y.~Wei, ``Compositional human pose
  regression,'' in {\em Proc. IEEE International Conference on Computer
  Vision}, pp.~2602--2611, 2017.

\bibitem{yang20183d}
W.~Yang, W.~Ouyang, X.~Wang, J.~Ren, H.~Li, and X.~Wang, ``{3D} human pose
  estimation in the wild by adversarial learning,'' in {\em Proc. IEEE
  Conference on Computer Vision and Pattern Recognition}, pp.~5255--5264, 2018.

\bibitem{chen2020towards}
Z.~Chen, Y.~Huang, H.~Yu, B.~Xue, K.~Han, Y.~Guo, and L.~Wang, ``Towards
  part-aware monocular {3D} human pose estimation: An architecture search
  approach,'' in {\em Proc. European Conference on Computer Vision},
  pp.~715--732, 2020.

\bibitem{lee2018propagating}
K.~Lee, I.~Lee, and S.~Lee, ``Propagating {LSTM}: {3D} pose estimation based on
  joint interdependency,'' in {\em Proc. European Conference on Computer
  Vision}, pp.~119--135, 2018.

\bibitem{chen20173d}
C.-H. Chen and D.~Ramanan, ``{3D} human pose estimation = {2D} pose estimation+
  matching,'' in {\em Proc. IEEE conference on Computer Vision and Pattern
  Recognition}, pp.~7035--7043, 2017.

\bibitem{tome2017lifting}
D.~Tome, C.~Russell, and L.~Agapito, ``Lifting from the deep: Convolutional
  {3D} pose estimation from a single image,'' in {\em Proc. IEEE Conference on
  Computer Vision and Pattern Recognition}, pp.~2500--2509, 2017.

\bibitem{tekin2017learning}
B.~Tekin, P.~M{\'a}rquez-Neila, M.~Salzmann, and P.~Fua, ``Learning to fuse
  {2D} and {3D} image cues for monocular body pose estimation,'' in {\em Proc.
  IEEE International Conference on Computer Vision}, pp.~3941--3950, 2017.

\bibitem{chen2018cascaded}
Y.~Chen, Z.~Wang, Y.~Peng, Z.~Zhang, G.~Yu, and J.~Sun, ``Cascaded pyramid
  network for multi-person pose estimation,'' in {\em Proc. IEEE Conference on
  Computer Vision and Pattern Recognition}, pp.~7103--7112, 2018.

\bibitem{sun2019deep}
K.~Sun, B.~Xiao, D.~Liu, and J.~Wang, ``Deep high-resolution representation
  learning for human pose estimation,'' in {\em Proc. Conference on Computer
  Vision and Pattern Recognition}, 2019.

\bibitem{Zheng2023Survey}
C.~Zheng, W.~Wu, C.~Chen, T.~Yang, S.~Zhu, J.~Shen, N.~Kehtarnavaz, and
  M.~Shah, ``Deep learning-based human pose estimation: A survey,'' {\em ACM
  Computing Surveys}, 2023.

\bibitem{zhao2019semantic}
L.~Zhao, X.~Peng, Y.~Tian, M.~Kapadia, and D.~N. Metaxas, ``Semantic graph
  convolutional networks for {3D} human pose regression,'' in {\em Proc. IEEE
  Conference on Computer Vision and Pattern Recognition}, pp.~3425--3435, 2019.

\bibitem{azizi20223d}
N.~Azizi, H.~Possegger, E.~Rodol{\`a}, and H.~Bischof, ``{3D} human pose
  estimation using {M}{\"o}bius graph convolutional networks,'' in {\em Proc.
  European Conference on Computer Vision}, pp.~160--178, 2022.

\bibitem{zhang2022group}
Z.~Zhang, ``Group graph convolutional networks for {3D} human pose
  estimation,'' in {\em Proc. British Machine Vision Conference}, 2022.

\bibitem{zhao2022graformer}
W.~Zhao, W.~Wang, and Y.~Tian, ``{GraFormer}: Graph-oriented transformer for
  {3D} pose estimation,'' in {\em Proc. IEEE Conference on Computer Vision and
  Pattern Recognition}, pp.~20438--20447, 2022.

\bibitem{PoseFormer:2021}
C.~Zheng, S.~Zhu, M.~Mendieta, T.~Yang, C.~Chen, and Z.~Ding, ``{3D} human pose
  estimation with spatial and temporal transformers,'' in {\em Proc. IEEE
  International Conference on Computer Vision}, 2021.

\bibitem{zou2020high}
Z.~Zou, K.~Liu, L.~Wang, and W.~Tang, ``High-order graph convolutional networks
  for {3D} human pose estimation,'' in {\em Proc. British Machine Vision
  Conference}, 2020.

\bibitem{quan2021higher}
J.~Quan and A.~{Ben Hamza}, ``Higher-order implicit fairing networks for {3D}
  human pose estimation,'' in {\em Proc. British Machine Vision Conference},
  2021.

\bibitem{Liu2020Disentangle}
Z.~Liu, H.~Zhang, Z.~Chen, Z.~Wang, and W.~Ouyang, ``Disentangling and unifying
  graph convolutions for skeleton-based action recognition,'' in {\em Proc.
  IEEE Conference on Computer Vision and Pattern Recognition}, pp.~143--152,
  2020.

\bibitem{Vaswani2017Tranformers}
A.~Vaswani, N.~Shazeer, N.~Parmar, J.~Uszkoreit, L.~Jones, A.~N. Gomez,
  L.~Kaiser, and I.~Polosukhin, ``Attention is all you need,'' in {\em Advances
  in Neural Information Processing Systems}, 2017.

\bibitem{pavllo20193d}
D.~Pavllo, C.~Feichtenhofer, D.~Grangier, and M.~Auli, ``{3D} human pose
  estimation in video with temporal convolutions and semi-supervised
  training,'' in {\em Proc. IEEE Conference on Computer Vision and Pattern
  Recognition}, pp.~7753--7762, 2019.

\bibitem{YujunCai:19}
Y.~Cai, L.~Ge, J.~Liu, J.~Cai, T.-J. Cham, J.~Yuan, and N.~M. Thalmann,
  ``Exploiting spatial-temporal relationships for 3d pose estimation via graph
  convolutional networks,'' in {\em Proc. IEEE Conference on Computer Vision
  and Pattern Recognition}, pp.~2272--2281, 2019.

\bibitem{zeng2020srnet}
A.~Zeng, X.~Sun, F.~Huang, M.~Liu, Q.~Xu, and S.~Lin, ``{SRNet}: Improving
  generalization in {3D} human pose estimation with a split-and-recombine
  approach,'' in {\em Proc. European Conference on Computer Vision},
  pp.~507--523, 2020.

\bibitem{liu2020attention}
R.~Liu, J.~Shen, H.~Wang, C.~Chen, S.-C. Cheung, and V.~Asari, ``Attention
  mechanism exploits temporal contexts: Real-time {3D} human pose
  reconstruction,'' in {\em Proc. IEEE Conference on Computer Vision and
  Pattern Recognition}, pp.~5064--5073, 2020.

\bibitem{chen2021anatomy}
T.~Chen, C.~Fang, X.~Shen, Y.~Zhu, Z.~Chen, and J.~Luo, ``Anatomy-aware {3D}
  human pose estimation with bone-based pose decomposition,'' {\em IEEE
  Transactions on Circuits and Systems for Video Technology}, vol.~32, no.~1,
  pp.~198--209, 2021.

\bibitem{cai2023htnet}
J.~Cai, H.~Liu, R.~Ding, W.~Li, J.~Wu, and M.~Ban, ``{HTNet:} human topology
  aware network for {3D} human pose estimation,'' in {\em Proc. IEEE
  International Conference on Acoustics, Speech and Signal Processing}, 2023.

\bibitem{martinez2017simple}
J.~Martinez, R.~Hossain, J.~Romero, and J.~J. Little, ``A simple yet effective
  baseline for {3D} human pose estimation,'' in {\em Proc. IEEE International
  Conference on Computer Vision}, pp.~2640--2649, 2017.

\bibitem{zou2021modulated}
Z.~Zou and W.~Tang, ``Modulated graph convolutional network for {3D} human pose
  estimation,'' in {\em Proc. IEEE International Conference on Computer
  Vision}, pp.~11477--11487, 2021.

\bibitem{multihop2022}
J.~Y. Lee and I.~G. Kim, ``Multi-hop modulated graph convolutional networks for
  {3D} human pose estimation,'' in {\em Proc. British Machine Vision
  Conference}, 2022.

\bibitem{Zaedul2023GSNet}
Z.~Islam and A.~{Ben Hamza}, ``Iterative graph filtering network for {3D} human
  pose estimation,'' {\em Journal of Visual Communication and Image
  Representation}, vol.~95, 2023.

\bibitem{Dosovitskiy2021VT}
A.~Dosovitskiy, L.~Beyer, A.~Kolesnikov, D.~Weissenborn, X.~Zhai,
  T.~Unterthiner, M.~Dehghani, M.~Minderer, G.~Heigold, S.~Gelly, J.~Uszkoreit,
  and N.~Houlsby, ``An image is worth 16x16 words: Transformers for image
  recognition at scale,'' in {\em International Conference on Learning
  Representations}, 2021.

\bibitem{Yu2016Multiscale}
F.~Yu and V.~Koltun, ``Multi-scale context aggregation by dilated
  convolutions,'' in {\em International Conference on Learning
  Representations}, 2016.

\bibitem{HuiZou:05}
H.~Zou and T.~Hastie, ``Regularization and variable selection via the elastic
  net,'' {\em Journal of the Royal Statistical Society. Series B}, vol.~60,
  no.~1, pp.~301--320, 2005.

\bibitem{ionescu2013human3}
C.~Ionescu, D.~Papava, V.~Olaru, and C.~Sminchisescu, ``{Human3.6M}: Large
  scale datasets and predictive methods for {3D} human sensing in natural
  environments,'' {\em IEEE Transactions on Pattern Analysis and Machine
  Intelligence}, vol.~36, no.~7, pp.~1325--1339, 2013.

\bibitem{mehta2017monocular}
D.~Mehta, H.~Rhodin, D.~Casas, P.~Fua, O.~Sotnychenko, W.~Xu, and C.~Theobalt,
  ``Monocular {3D} human pose estimation in the wild using improved cnn
  supervision,'' in {\em Proc. International Conference on 3D Vision},
  pp.~506--516, 2017.

\bibitem{pavlakos2018ordinal}
G.~Pavlakos, X.~Zhou, and K.~Daniilidis, ``Ordinal depth supervision for {3D}
  human pose estimation,'' in {\em Proc. IEEE Conference on Computer Vision and
  Pattern Recognition}, pp.~7307--7316, 2018.

\bibitem{liu2021graph}
J.~Liu, J.~Rojas, Y.~Li, Z.~Liang, Y.~Guan, N.~Xi, and H.~Zhu, ``A graph
  attention spatio-temporal convolutional network for {3D} human pose
  estimation in video,'' in {\em Proc. IEEE International Conference on
  Robotics and Automation}, pp.~3374--3380, 2021.

\bibitem{zeng2021learning}
A.~Zeng, X.~Sun, L.~Yang, N.~Zhao, M.~Liu, and Q.~Xu, ``Learning skeletal graph
  neural networks for hard {3D} pose estimation,'' in {\em Proc. IEEE
  International Conference on Computer Vision}, pp.~11436--11445, 2021.

\bibitem{Zhao2023PoseFormerV2}
Q.~Zhao, C.~Zheng, M.~Liu, P.~Wang, and C.~Chen, ``{PoseFormerV2}: Exploring
  frequency domain for efficient and robust {3D} human pose estimation,'' in
  {\em Proc. IEEE Conference on Computer Vision and Pattern Recognition}, 2023.

\bibitem{hossain2018exploiting}
M.~R.~I. Hossain and J.~J. Little, ``Exploiting temporal information for {3D}
  human pose estimation,'' in {\em Proc. European Conference on Computer
  Vision}, pp.~68--84, 2018.

\bibitem{lin2019trajectory}
J.~Lin and G.~H. Lee, ``Trajectory space factorization for deep video-based
  {3D} human pose estimation,'' in {\em Proc. British Machine Vision
  Conference}, 2019.

\bibitem{ChenLiLee:2020}
C.~Li and G.~H. Lee, ``Weakly supervised generative network for multiple {3D}
  human pose hypotheses,'' in {\em Proc. British Machine Vision Conference},
  2020.

\bibitem{li2019generating}
C.~Li and G.~H. Lee, ``Generating multiple hypotheses for {3D} human pose
  estimation with mixture density network,'' in {\em Proc. IEEE Conference on
  Computer Vision and Pattern Recognition}, pp.~9887--9895, 2019.

\bibitem{Habibie:19}
I.~Habibie, W.~Xu, D.~Mehta, G.~Pons-Moll, and C.~Theobalt, ``In the wild human
  pose estimation using explicit {2D} features and intermediate {3D}
  representations,'' in {\em Proc. IEEE Conference on Computer Vision and
  Pattern Recognition}, pp.~10905--10914, 2019.

\bibitem{xu2021graph}
T.~Xu and W.~Takano, ``Graph stacked hourglass networks for {3D} human pose
  estimation,'' in {\em Proc. IEEE Conference on Computer Vision and Pattern
  Recognition}, pp.~16105--16114, 2021.

\bibitem{Zhan_2022_CVPR}
Y.~Zhan, F.~Li, R.~Weng, and W.~Choi, ``{Ray3D}: Ray-based {3D} human pose
  estimation for monocular absolute {3D} localization,'' in {\em Proc. IEEE
  Conference on Computer Vision and Pattern Recognition}, pp.~13116--13125,
  2022.

\bibitem{hassanin2022crossformer}
M.~Hassanin, A.~Khamiss, M.~Bennamoun, F.~Boussaid, and I.~Radwan,
  ``Crossformer: Cross spatio-temporal transformer for {3D} human pose
  estimation,'' {\em arXiv preprint arXiv:2203.13387}, 2022.

\bibitem{Mao2019DCT}
W.~Mao, M.~Liu, M.~Salzmann, and H.~Li, ``Learning trajectory dependencies for
  human motion prediction,'' in {\em Proc. IEEE International Conference on
  Computer Vision}, pp.~9489--9497, 2019.

\end{thebibliography}

\end{document}